\documentclass{article}
\usepackage{iclr2026_conference}

\usepackage{microtype}
\usepackage{graphicx}
\usepackage{booktabs}
\usepackage{longtable}
\usepackage{multirow}
\usepackage{array}
\usepackage{mathtools}
\usepackage{amssymb}
\usepackage[table]{xcolor}
\usepackage{enumitem}
\usepackage{etoc}
\usepackage[font=small,labelfont=bf]{caption}
\usepackage{subcaption}
\usepackage{wrapfig}
\usepackage{placeins}
\usepackage[misc]{ifsym}
\usepackage{simpleicons}
\usepackage{twemojis}
\usepackage{timelens_report}

\usepackage{amsmath,amsfonts,bm}

\def\eqref#1{equation~\ref{#1}}

\def\1{\bm{1}}

\DeclareMathAlphabet{\mathsfit}{\encodingdefault}{\sfdefault}{m}{sl}
\SetMathAlphabet{\mathsfit}{bold}{\encodingdefault}{\sfdefault}{bx}{n}

\usepackage{hyperref}
\usepackage[nameinlink,capitalize,noabbrev]{cleveref}
\usepackage{url}
\usepackage{xspace}

\hypersetup{
    colorlinks=true,
    linkcolor=linkblue,
    citecolor=citegreen,
    urlcolor=urlmagenta,
    pdftitle={TimeLens2: Generalist Video Temporal Grounding with Multimodal LLMs},
    pdfauthor={Yuhan Zhu, Changlian Ma, Xinhao Li, Xiangyu Zeng, Zhiqiu Zhang, Yuandong Yang, Tianxiang Jiang, Songze Li, Jun Zhang, Xinyu Chen, Haoran Chen, Zikang Wang, Limin Wang}
}

\graphicspath{{figures/}}

\renewcommand{\authornamesize}{\fontsize{10pt}{10pt}\selectfont}
\renewcommand{\reportshorttitle}{TimeLens2}
\newcommand{\affmark}[1]{\textsuperscript{\normalfont #1}}
\newcommand{\nju}{\affmark{1}}
\newcommand{\ailab}{\affmark{2}}
\newcommand{\sjtu}{\affmark{3}}
\newcommand{\zju}{\affmark{4}}
\newcommand{\ustc}{\affmark{5}}
\newcommand{\fudan}{\affmark{6}}
\newcommand{\corrauth}{\Letter}

\title{TimeLens2: Generalist Video Temporal Grounding with Multimodal LLMs}

\author{
\begin{minipage}{\textwidth}
\raggedright
Yuhan Zhu\affmark{* 1, 2}\hspace{0.5em}
Changlian Ma\affmark{* 1, 2}\hspace{0.5em}
Xiangyu Zeng\affmark{* 1, 2}\hspace{0.5em}
Xinhao Li\affmark{* 1}\hspace{0.5em}
Zhiqiu Zhang\affmark{* 3, 2} \\ \vspace{0.45em}
Songze Li\affmark{6, 2}\hspace{0.5em}
Jun Zhang\affmark{1}\hspace{0.5em} 
Tianxiang Jiang\affmark{5, 2}\hspace{0.5em}
Yuandong Yang\affmark{1}\hspace{0.5em}
Ziang Yan\affmark{4, 2}\hspace{0.5em} \\ \vspace{0.45em}
Zikang Wang\affmark{3, 2}\hspace{0.5em} 
Xinyu Chen\affmark{1, 2}\hspace{0.5em} 
Haoran Chen\affmark{1, 2}\hspace{0.5em} 
Shaowei Zhang\affmark{3, 2}\hspace{0.5em}
Limin Wang\affmark{1, 2, \corrauth} \\
\vspace{0.8em}
{\normalfont\fontsize{8.5pt}{10pt}\selectfont
\nju Nanjing University \quad
\ailab Shanghai AI Laboratory \quad
\sjtu Shanghai Jiao Tong University \\ \vspace{0.2em}
\zju Zhejiang University \quad
\ustc University of Science and Technology of China \quad
\fudan Fudan University}
\end{minipage}
}

\newcommand{\methodname}{TimeLens2\xspace}

\iclrfinalcopy
\begin{document}

\maketitle
\setupreportpagestyle
\thispagestyle{reportfirst}

\begingroup
\renewcommand{\thefootnote}{\fnsymbol{footnote}}
\footnotetext[1]{~Equal contribution.}
\endgroup
\begingroup
\renewcommand{\thefootnote}{\Letter}
\footnotetext[1]{~Corresponding author: \texttt{lmwang@nju.edu.cn}.}
\endgroup

\begin{abstract}
Video multimodal large language models (MLLMs) can describe what happens in a video, but rarely identify when the supporting evidence occurs. We study generalist video temporal grounding, in which one model predicts a variable-cardinality set of evidence intervals across video lengths, domains, query forms, and viewpoints. Existing training strategies are misaligned with this set-valued task: long-video labels often rely on brittle one-pass annotation, while reinforcement-learning rewards either fail to distinguish non-overlapping predictions or require fragile segment matching. \methodname treats temporal evidence as an interval set throughout supervision and optimization. TimeLens2-93K constructs reliable multi-span supervision through caption-derived proposals, independent localization, cross-agent consensus, semantic verification, and boundary refinement. Our temporal Wasserstein reward computes exact one-dimensional \(W_1\) between uniform distributions over merged interval supports, providing dense, matching-free feedback under unequal cardinalities and equivalent fragmentation; temporal IoU complements it with precise-overlap feedback. Across seven benchmarks, \methodname-2B outperforms all size-matched baselines on every benchmark, while the 4B and 8B variants achieve state-of-the-art performance, surpassing open-source models with up to 397B parameters. The 2B, 4B, and 8B variants improve over their Qwen3-VL backbones by 14.2, 13.0, and 18.1 mIoU points, respectively.

\par\vspace{0.55em}
\begingroup
\setlength{\tabcolsep}{0pt}
\renewcommand{\arraystretch}{1.18}
\small
\noindent\begin{tabular}{@{}c@{\hspace{0.65em}}l@{}}
\raisebox{-0.42ex}{\twemoji[height=1.08em]{house}} &
\href{https://mcg-nju.github.io/TimeLens2}%
{{\color{reportblue}\nolinkurl{https://mcg-nju.github.io/TimeLens2}}} \\
\raisebox{-0.44ex}{{\fontsize{9pt}{9pt}\selectfont\color{reportink}\simpleicon{github}}} &
\href{https://github.com/MCG-NJU/TimeLens2}%
{{\color{reportblue}\nolinkurl{https://github.com/MCG-NJU/TimeLens2}}} \\
\raisebox{-0.42ex}{\twemoji[height=1.08em]{hugs}} &
\href{https://huggingface.co/collections/MCG-NJU/timelens2}%
{{\color{reportblue}\nolinkurl{https://huggingface.co/collections/MCG-NJU/timelens2}}}
\end{tabular}
\endgroup

\end{abstract}

\section{Introduction}
\label{sec:introduction}

Video multimodal large language models (MLLMs) promise to make growing video archives searchable through language~\citep{videochatgpt,videollama,videochat-flash,qwen3vl,Video-o3}. Yet answering \emph{what} happened is not enough: users must also know \emph{when} the supporting evidence appears. Without temporal support, users must still search the full timeline, and even correct descriptions remain unverifiable. Temporal grounding is therefore the video analogue of citation: it makes outputs traceable by requiring precise, potentially disjoint evidence intervals rather than only fluent responses~\citep{Charades,ActivityNet,Ren2023TimeChat,huang2024vtimellm,wang2024groundedvideollm}.

Temporal grounding is not a single-regime task: supporting evidence may occupy only seconds of an hour-long video, recur at disjoint moments, and appear in third- or first-person footage, while queries may be descriptions or questions. We therefore study \emph{generalist temporal grounding}, where a single MLLM with a unified output interface localizes single- or multi-interval evidence across video lengths, domains, query forms, and viewpoints.

\begin{figure}[t]
    \centering
    \includegraphics[width=\linewidth]{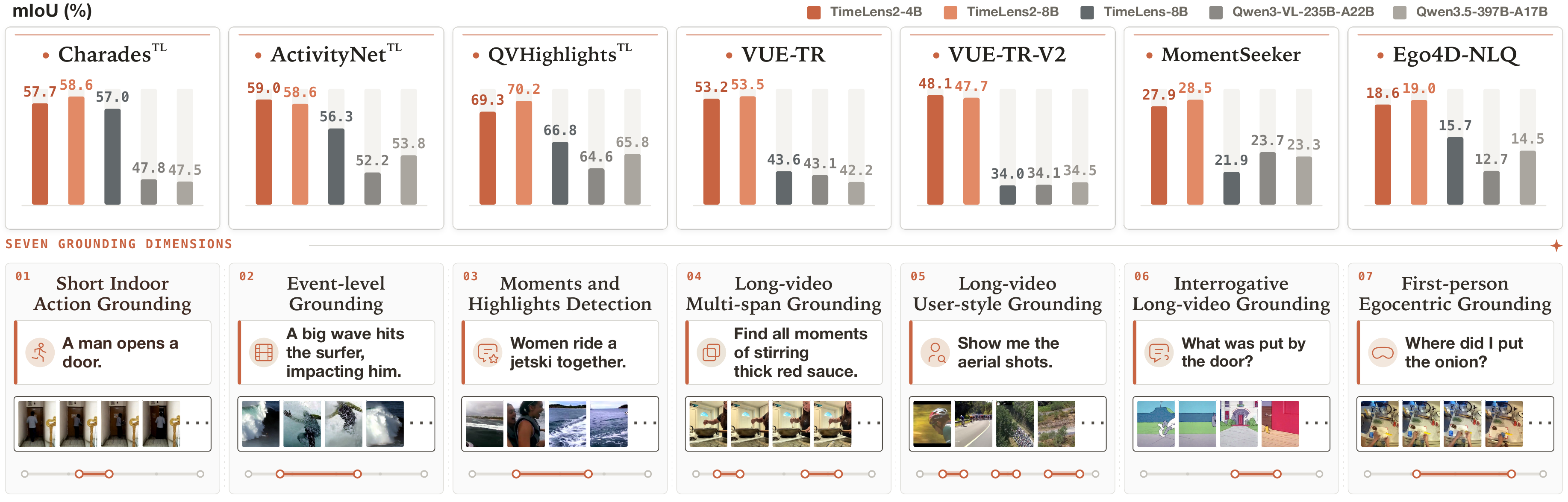}
    \vspace{-0.49cm}
    \caption{\textbf{\methodname as a generalist video temporal-grounding model.} \textbf{Top:} mIoU across seven benchmarks covering diverse evaluation dimensions. \textbf{Bottom:} Representative queries from each benchmark.}
    \label{fig:teaser}
\end{figure}

Two structural mismatches stand in the way. The first is in supervision. Temporal evidence is a set of intervals, yet long-video labels     are often produced as a single global annotation decision. Because relevant moments are sparse and similar distractors dominate the timeline, a one-pass annotator may confuse similar occurrences, miss repeated evidence, or produce imprecise boundaries~\citep{qian2024momentor,timesuite,zhang2026timelens}. Meanwhile, many existing datasets emphasize short videos, declarative captions, and single-span answers~\citep{Charades,ActivityNet,hendricks2017didemo}. Long-video annotation is therefore not merely a scaling problem; it is an evidence-verification problem.

Even with reliable supervision, optimization remains mismatched with temporal evidence: next-token prediction learns timestamp generation effectively, but lacks an explicit interval-level objective~\citep{Ren2023TimeChat,huang2024vtimellm,timesuite,zhang2026timelens}. Reinforcement learning with temporal IoU (tIoU) better aligns training with evaluation~\citep{Time-R1,VideoChat-R1,li2025rlt,chen2025vtgrl}, but assigns zero to all disjoint predictions, leaving near misses indistinguishable from distant errors. For multi-span targets, one-to-one matching~\citep{MUSEG} is also fragile under fragmentation or unequal cardinality.

We introduce \methodname, a generalist temporal-grounding MLLM that treats evidence as a first-class interval set throughout training. The central idea is to make evidence \emph{verifiable} when constructing supervision and \emph{geometry-aware} when optimizing predictions.

TimeLens2-93K turns one brittle global annotation decision into a sequence of increasingly focused ones. Hierarchical, time-stamped captions first provide long-range context for proposing declarative queries and coarse single- or multi-span evidence; complementary grounding agents then independently relocalize that evidence from the video. Temporal consensus and semantic verification remove unstable or mismatched instances, after which local refinement sharpens only the surviving boundaries. By progressively narrowing both the search space and the error type, the pipeline preserves repeated evidence while making long-video supervision more reliable.

Once reliable interval labels are available, training separates capability acquisition from geometric calibration. Long-context supervision first teaches the model to search full videos and generate variable-cardinality interval sets; GRPO then directly calibrates where those intervals lie~\citep{shao2024deepseekmath}. tIoU measures how much predicted support is already correct, but not how a disjoint prediction should move. Our temporal Wasserstein reward supplies this missing geometry by computing the exact one-dimensional \(W_1\) distance between uniform distributions over merged predicted and target supports, providing dense, matching-free guidance for near misses and unequal-cardinality outputs while remaining invariant to equivalent fragmentation.

\Cref{fig:teaser} summarizes performance across seven benchmarks: \methodname-2B, \methodname-4B, and \methodname-8B reach 44.5, 47.7, and 48.0 average mIoU, respectively. The 2B, 4B, and 8B variants improve over their Qwen3-VL backbones by 14.2, 13.0, and 18.1 points~\citep{qwen3vl}; the 4B model also surpasses Qwen3.5-397B-A17B~\citep{qwen35blog} on every benchmark by 7.5 points on average. In data ablations, declarative-only TimeLens2-93K raises question-form grounding on MomentSeeker from 15.3 to 25.8 mIoU~\citep{momentseeker}. Temporal Wasserstein restores informative preferences for 75.8\% of all-zero-tIoU GRPO groups.

Our contributions are threefold:
\begin{itemize}[leftmargin=*]
    \item We introduce TimeLens2-93K, a staged evidence-verification pipeline that produces reliable single- and multi-span temporal grounding supervision for long videos.
    \item We propose a matching-free temporal Wasserstein reward based on exact one-dimensional \(W_1\) over merged interval support, providing graded, fragmentation-invariant feedback for disjoint and unequal-cardinality predictions.
    \item We demonstrate compact 2B, 4B, and 8B generalist models across seven benchmarks covering long-video, multi-span, question-form, and egocentric grounding.
\end{itemize}

\section{Related Work}
\label{sec:related-work}

\paragraph{Temporal Grounding Models.}
Video temporal grounding has evolved from dedicated localization architectures to generative MLLMs. Classical methods match moment proposals or regress boundaries~\citep{Charades,zhu2024dualdetrsmultilabeltemporal}; recent systems instead express temporal evidence through a language interface. TimeChat and VTimeLLM introduce temporal instruction tuning and boundary-aware training~\citep{Ren2023TimeChat,huang2024vtimellm}, while LITA and Grounded-VideoLLM develop timestamp-aware representations~\citep{huang2024lita,wang2024groundedvideollm}. TimeSuite features grounding-centric instruction tuning to strengthen both long-video QA and temporal localization~\citep{timesuite}, whereas TimeLens systematically studies data quality and training recipes~\citep{zhang2026timelens}. In parallel, general video representations and embeddings broaden transferable video evidence search~\citep{wang2026internvideonextgeneralvideofoundation,liu2025reasoningguidedembeddingsleveraging,zhu2026freeretmllmstrainingfreeretrievers}. This shift makes grounding a general MLLM capability. Yet supervision has not kept pace: transfer across video lengths, evidence cardinalities, viewpoints, and query forms remains unresolved. \methodname is designed around this broader regime.

\paragraph{Temporal Grounding Data.}
The data landscape reflects the same progression. Early benchmarks primarily pair short or domain-specific videos with a single descriptive moment~\citep{regneri2013tacos,hendricks2017didemo,Charades,ActivityNet}. Later datasets expand individual axes of difficulty, including subtitle-aware retrieval, repeated evidence, long-form video, and egocentric search~\citep{lei2020tvr,QVHighlights,soldan2022mad,grauman2022ego4d}. MLLM recipes then convert timestamped annotations into instruction-following conversations~\citep{Ren2023TimeChat,huang2024vtimellm,timesuite,qian2024momentor}, often scaling through dense captions, procedural videos, or weak narration alignment~\citep{zhou2018youcook2,miech2019howto100m,tang2019coin,zhukov2019crosstask}. Scale, however, does not by itself resolve supervision quality: in long videos, coarse alignment readily becomes a wrong occurrence, a missed repeat, or an imprecise boundary. TimeLens2-93K targets this bottleneck with consensus-based localization and semantic verification, yielding reliable single- and multi-span supervision across long, diverse videos.

\paragraph{Temporal Grounding Optimization.}
A parallel mismatch appears in optimization. Supervised instruction tuning teaches the syntax of timestamps through token likelihood, not the quality of the localized evidence. GRPO/RLVR methods make grounding verifiable through format and temporal-overlap rewards, and subsequent work extends this recipe with improved reward design, data selection, refusal, curricula, and multi-segment reasoning~\citep{shao2024deepseekmath,Time-R1,VideoChat-R1,li2025rlt,chen2025vtgrl,yue2025tempor0,dong2025videotgr1,wu2025tempr1,MUSEG,li2026videoopd}. Nevertheless, overlap supplies geometry only after prediction and target intersect. MUSEG introduces pairwise NGIoU for disjoint segments, but its one-to-one matching remains sensitive to fragmentation, merges, and unequal cardinality~\citep{MUSEG}. The issue is therefore not merely reward sparsity; it is also the choice of representation for set-valued evidence. Wasserstein geometry provides a useful precedent: NWD represents spatial boxes by Gaussian surrogates and applies pairwise \(W_2\) in supervised tiny-object detection~\citep{NWD}. Our objective instead treats the merged support of a variable-cardinality interval set as a one-dimensional distribution and computes its exact \(W_1\) distance.

\section{TimeLens2}
\label{sec:method}

\subsection{TimeLens2-93K: Scalable Temporal Grounding Data Construction}
\label{subsec:data-construction}

Long videos pose a fundamental supervision challenge: full-video context is needed to teach evidence search, yet sparse evidence and growing distractors make precise annotation unreliable. We construct TimeLens2-93K from long, diverse web videos and represent each example as \((v,q,\mathcal{Y})\), where \(\mathcal{Y}=\{[s_k,e_k]\}_{k=1}^{K}\) is a variable-cardinality set of supporting intervals. This formulation unifies single- and multi-span grounding while preserving the context in which the evidence must be found.

\begin{figure}[t]
    \centering
    \includegraphics[width=\textwidth]{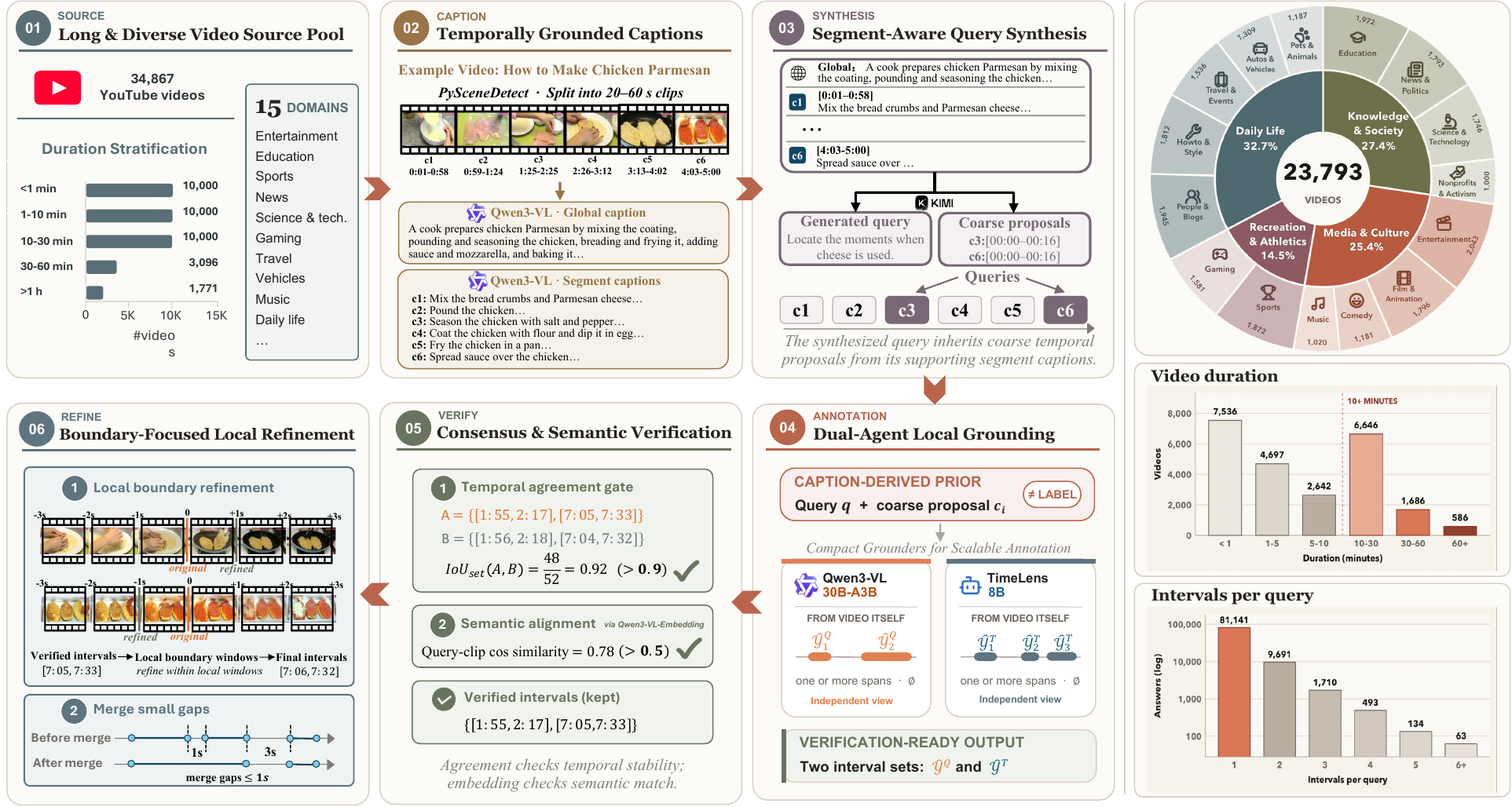}
    \caption{\textbf{TimeLens2-93K construction pipeline and corpus profile.} Steps 1--3 construct caption-derived queries and coarse proposals; Steps 4--6 relocalize proposals with independent agents, verify interval sets, and refine their boundaries. The right column summarizes the retained corpus.}
    \label{fig:data-pipeline}
\end{figure}

As summarized in \Cref{fig:data-pipeline}, our pipeline separates candidate construction (Steps 1--3) from label determination (Steps 4--6). Hierarchical, time-stamped captions yield declarative queries and coarse single- or multi-span proposals; independent agents then relocalize the proposed evidence directly from short video clips. Temporal consensus and semantic verification reject unstable or mismatched instances, after which local refinement sharpens only the surviving boundaries. This staged factorization lets the model learn from full-video context while producing labels through controlled local decisions. The resulting corpus contains 23,793 videos and 93,232 grounding instances, including 12,091 instances with multiple supporting intervals.

\paragraph{Long and diverse video source pool.}
To make temporal search difficulty a deliberate design axis rather than an incidental consequence of web sampling, we stratify 34,867 YouTube videos across five duration ranges, from under one minute to over one hour, and diverse visual domains. Duration controls the search horizon and distractor load, while domain breadth diversifies the evidence to be grounded.

\paragraph{Hierarchical temporally grounded captions.}
Directly generating queries and localizing their evidence over a long video would conflate video understanding, query synthesis, and localization in one brittle step. We therefore convert each video into a hierarchical, time-stamped description that serves as a semantic index for subsequent query and proposal generation. A global caption captures subjects, setting, actions, and temporal progression to disambiguate recurring events; segment captions enumerate visually verifiable actions, object states, and interactions within bounded intervals. We construct this index by partitioning \(v\) into semantically coherent clips \(\mathcal{C}(v)=\{c_i=[t_i,t_{i+1}]\}_{i=1}^{N}\) using content-based PySceneDetect boundaries constrained to 20--60 seconds. Compared with uniform slicing, these boundaries better preserve visual continuity while keeping clips short enough for reliable captioning. Qwen3-VL-235B-A22B~\citep{qwen3vl} generates both caption levels, emphasizing observable events and temporal order.

\paragraph{Segment-aware query synthesis.}
Query synthesis and evidence proposal are naturally coupled: once a query is derived from time-stamped segment descriptions, its candidate support is already encoded in the same hierarchy. Rather than discard this alignment and recover it through a separate full-video retrieval stage, we condition Kimi-K2.5~\citep{Kimi-K2.5} on the global and segment captions to jointly generate a non-redundant declarative query and select all segments whose captions support it. These segments form a coarse proposal \(\mathcal{P}(q)\subseteq \mathcal{C}(v)\), which may include disjoint clips when evidence recurs. Caption-derived proposals provide temporal priors for downstream agent annotation.

\paragraph{Dual-agent local grounding.}
Caption-derived proposals narrow the search space but are not labels: captions may contain unsupported content, and segment boundaries rarely match true event boundaries. We therefore relocalize each proposal from the video itself. For each query-proposal pair \((q,c_i)\), Qwen3-VL-30B-A3B~\cite{qwen3vl} and TimeLens-8B~\citep{zhang2026timelens} independently return one or more intervals at one-second resolution or an empty set. Both are strong yet efficient, while their distinct inductive biases provide two views for consensus. Empty outputs reject unsupported proposals; multi-interval outputs preserve repeated evidence within a clip. We map clip-relative timestamps to the original timeline and aggregate them into \(\hat{\mathcal{Y}}^{\mathrm{Q}}\) and \(\hat{\mathcal{Y}}^{\mathrm{T}}\). This converts a coarse text match into a verifiable visual decision with explicit boundaries.

\paragraph{Cross-agent consensus and semantic verification.}
Independent localization makes temporal labels testable: reliable evidence should be recovered across annotators with different inductive biases. Yet agreement alone cannot rule out a shared semantic error. We therefore verify two distinct properties: temporal reproducibility and query--evidence alignment. For temporal consensus, let \(\operatorname{merge}(\cdot)\) denote interval union and \(|\cdot|\) total duration. Comparing merged supports for multi-span outputs:
\[
\operatorname{IoU}_{\mathrm{set}}(\mathcal{A},\mathcal{B})
=
\frac{|\operatorname{merge}(\mathcal{A}) \cap \operatorname{merge}(\mathcal{B})|}
{|\operatorname{merge}(\mathcal{A}) \cup \operatorname{merge}(\mathcal{B})|}.
\]
We retain instances with \(\operatorname{IoU}_{\mathrm{set}}(\hat{\mathcal{Y}}^{\mathrm{Q}},\hat{\mathcal{Y}}^{\mathrm{T}})>0.9\) and keep the Qwen annotation as canonical; post-hoc fusion could create boundaries predicted by neither model. To verify semantic validity, we encode each retained target clip and its query with Qwen3-VL-Embedding~\cite{Qwen3-VL-Embedding} and require a normalized text--video cosine similarity of at least 0.5. Thus, consensus tests whether a localization is reproducible, while embedding verification tests whether the reproduced evidence is relevant.

\paragraph{Boundary-focused local refinement.}
Once consensus and semantic verification establish the event identity, the remaining uncertainty lies at its visual transitions. We therefore cast refinement as local change-point detection: for each retained boundary, Qwen3-VL-235B-A22B observes a \(\pm3\)-second neighborhood and predicts a refined transition point. For multi-interval labels, we merge adjacent spans separated by at most one second, treating such gaps as boundary jitter or brief occlusion rather than semantic breaks. Restricting refinement to verified local windows keeps this stronger model practical at scale.

The resulting cascade uses global context to propose evidence and concentrated local computation to verify and refine high-confidence interval-set labels.

\subsection{TimeLens2 Model: Learning Generalist Video Temporal Grounding}
\label{subsec:model}

Verified interval labels improve supervision, but a generative MLLM must still learn to search long contexts and express interval sets before being optimized by interval-level metrics. We therefore separate capability acquisition from geometric calibration: long-context supervised training learns evidence search and output conventions, while reinforcement learning directly optimizes decoded interval sets against non-differentiable temporal objectives.

\paragraph{Long-context supervised grounding.}
Temporal grounding is a search problem before it is a boundary-estimation problem: target-centered training removes the distractors that require evidence selection. We therefore fine-tune Qwen3-VL~\citep{qwen3vl} on long-context examples drawn primarily from TimeLens2-93K and TimeLens-100K~\citep{zhang2026timelens}. All samples use \((v,q,\mathcal{Y})\), where \(\mathcal{Y}\) contains one or more supporting intervals; the official Ego4D-NLQ training split~\cite{grauman2022ego4d} is added to broaden supervision to first-person video. Retaining full-video context forces the model to isolate sparse evidence among competing events, jointly learning evidence search and interval-set generation.

\paragraph{Instruction and response-format diversity.}
Temporal grounding should be invariant to how evidence is requested and serialized. A single prompt and timestamp format can entangle localization with surface-form imitation, limiting cross-benchmark transfer. For each training example, we independently sample the grounding instruction, answer syntax, and timestamp encoding. The same \(\mathcal{Y}\) is rendered in diverse single- or multi-span formats, including JSON, natural language, key--value or range styles, and alternative timestamp conventions. This encourages protocol-invariant interval semantics without synthetic paraphrases that might shift the evidence target.

\paragraph{Rollout-guided hard-sample mining.}
After SFT, GRPO should focus on what the current policy still fails to localize. Uniform prompt sampling wastes updates on solved examples, while static difficulty heuristics may not reflect model-specific failures. We therefore generate multiple off-policy completions from the SFT checkpoint for each example in TimeLens2-93K and TimeLens-100K, using their mean tIoU as an empirical estimate of model competence. Examples with lower scores receive higher sampling weights during GRPO~\citep{shao2024deepseekmath}. This converts the model's own failure distribution into an adaptive curriculum, concentrating optimization on unresolved localization errors rather than mirroring the dataset distribution.

\begin{figure}[t]
    \centering
    \includegraphics[width=\textwidth]{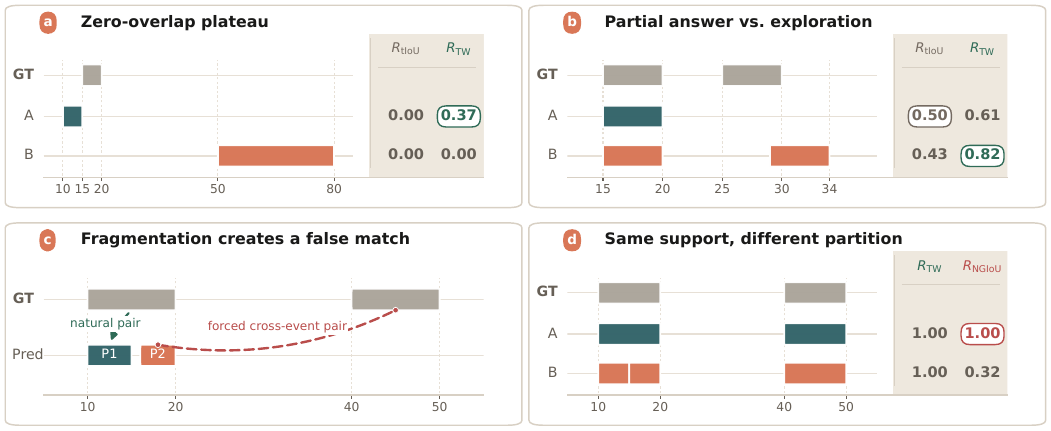}
    \caption{\textbf{Overlap plateaus and matching artifacts in temporal interval-set rewards.} \textbf{(a)} tIoU ties the near miss A and distant error B at zero, whereas \(R_{\mathrm{TW}}\) ranks them by temporal proximity. \textbf{(b)} tIoU favors the single-moment prediction A, while \(R_{\mathrm{TW}}\) favors B, which also recovers part of the second target moment. MUSEG addresses these failures with NGIoU under one-to-one matching~\cite{MUSEG}, but introduces new artifacts: \textbf{(c)} fragmentation forces P2 into a false cross-event match; \textbf{(d)} identical merged support receives different \(R_{\mathrm{NGIoU}}\) scores solely because the prediction is partitioned differently.}
    \label{fig:tw-reward-motivation}
\end{figure}

\paragraph{Temporal Wasserstein reward.}
An effective RL reward must rank imperfect localizations by their progress toward the target, not merely recognize existing overlap. Yet tIoU, the dominant reward in prior video temporal grounding RL~\cite{Time-R1,zhang2026timelens,VideoChat-R1}, is better suited to evaluation than reward shaping. In \Cref{fig:tw-reward-motivation}(a), it assigns zero to both the near miss A and the distant error B, discarding their temporal geometry. In the multi-moment example of \Cref{fig:tw-reward-motivation}(b), tIoU favors the partial answer A over B (0.50 vs.\ 0.43), even though B covers both target moments. When a GRPO group falls on such a plateau, tIoU provides no relative learning signal after mean centering. We therefore complement it with a temporal Wasserstein reward, \(R_{\mathrm{TW}}\), which ranks A above B in (a) and B above A in (b) by measuring the transport needed to align predicted temporal mass with the target support.

Accordingly, each sampled completion is parsed into a predicted interval set \(\hat{\mathcal{Y}}\) and scored by three complementary terms:
\[
R(\hat{\mathcal{Y}},\mathcal{Y})
= R_{\mathrm{tIoU}}(\hat{\mathcal{Y}},\mathcal{Y})
+ R_{\mathrm{TW}}(\hat{\mathcal{Y}},\mathcal{Y})
- \mathbf{1}_{\mathrm{invalid}} .
\]
The invalid-output indicator penalizes responses that cannot be parsed into valid intervals. For valid predictions, set-level tIoU rewards the target support already recovered:
\[
R_{\mathrm{tIoU}}(\hat{\mathcal{Y}},\mathcal{Y})
=
\frac{|\operatorname{merge}(\hat{\mathcal{Y}}) \cap \operatorname{merge}(\mathcal{Y})|}
{|\operatorname{merge}(\hat{\mathcal{Y}}) \cup \operatorname{merge}(\mathcal{Y})|}.
\]

tIoU measures how much support is already correct, but not how an incorrect prediction should move. To expose this missing geometry, we lift each non-empty interval set \(\mathcal{A}\) to a uniform temporal distribution over its merged support:
\[
\mu_{\mathcal{A}}(t)
=
\frac{\mathbf{1}\{t \in \operatorname{merge}(\mathcal{A})\}}
{\ell(\mathcal{A})}.
\]
Here \(\ell(\mathcal{A})=|\operatorname{merge}(\mathcal{A})|\). This representation makes the reward depend on where evidence lies, rather than how the same support is partitioned into spans. The 1-Wasserstein distance measures the transport needed to align the predicted and target mass; in one-dimensional time, it has the exact CDF form
\[
W(\hat{\mathcal{Y}},\mathcal{Y})
=
W_1(\mu_{\hat{\mathcal{Y}}}, \mu_{\mathcal{Y}})
=
\int_{\mathbb{R}}
\left|
F_{\hat{\mathcal{Y}}}(t) - F_{\mathcal{Y}}(t)
\right| \, dt,
\]
where \(F_{\mathcal{A}}(t)=\int_{-\infty}^{t}\mu_{\mathcal{A}}(\tau)\,d\tau\). We convert the distance into a scale-normalized similarity:
\[
R_{\mathrm{TW}}(\hat{\mathcal{Y}},\mathcal{Y})
=
\exp\left(
-
\frac{W(\hat{\mathcal{Y}},\mathcal{Y})}
{|\operatorname{merge}(\mathcal{Y})| + \epsilon}
\right),
\]
where the target-duration denominator makes the score comparable across target event scales without allowing overly broad predictions to enlarge their own normalization factor.

\paragraph{Discussion.} Boundary- or center-based losses can supply the geometry missing from tIoU for a single interval, but multi-span grounding additionally requires correspondences between predicted and target intervals. MUSEG handles this with one-to-one matching followed by pairwise NGIoU~\cite{MUSEG}; the matching itself, however, is fragile. Fragmentation sends P2 to the wrong target in \Cref{fig:tw-reward-motivation}(c), while predictions with identical merged support receive NGIoU scores of 1.0 and 0.32 solely because they are partitioned differently in (d). \(R_{\mathrm{TW}}\) instead compares merged temporal mass, avoiding correspondence and providing dense, partition-invariant credit. Thus, tIoU anchors support overlap, \(R_{\mathrm{TW}}\) supplies geometry, and the parse penalty enforces valid outputs.

\section{Experiments}
\label{sec:experiments}

\subsection{Experimental Setup}
\label{subsec:temporal-grounding-benchmarks}

\paragraph{Benchmarks.}
We evaluate \methodname on the seven temporal grounding benchmarks: the TimeLens-Bench~\citep{zhang2026timelens} re-annotations of Charades-STA~\citep{Charades}, ActivityNet Captions~\citep{ActivityNet}, and QVHighlights~\citep{QVHighlights}; the vision subsets of the long-video VUE-TR~\citep{vidi} and VUE-TR-V2~\citep{vidi2}; the text-only query subset of MomentSeeker~\citep{momentseeker} for question-form localization; and the official validation split of the egocentric Ego4D-NLQ~\citep{grauman2022ego4d}.

\paragraph{Evaluation metrics.}
We report mean temporal Intersection-over-Union (mIoU) and Recall@1 at tIoU thresholds of 0.3, 0.5, and 0.7. mIoU measures average boundary overlap, whereas R1@threshold is the fraction of top predictions meeting the specified overlap. The main paper reports mIoU and R1@0.5; R1@0.3 and R1@0.7 are deferred to the supplementary material~(\Cref{tab:temporal-grounding-benchmarks-full}).

\paragraph{Implementation details.}
Unless otherwise specified, \methodname uses Qwen3-VL-2B-Instruct, Qwen3-VL-4B-Instruct, or Qwen3-VL-8B-Instruct~\citep{qwen3vl}. Following the two-stage recipe in \Cref{subsec:model}, SFT uses TimeLens2-93K, TimeLens-100K~\citep{zhang2026timelens}, and the official Ego4D-NLQ-v2 training split~\citep{grauman2022ego4d}, whose answer windows are converted to the shared \((v,q,\mathcal{Y})\) format. For the 4B and 8B variants, we train for one epoch on packed sequences of up to 100K tokens, with a global batch size of 256 and a learning rate of \(5\times10^{-6}\) under cosine decay. For the 2B variant, we instead use a 160K-token packed length, a global batch size of 128, and a learning rate of \(10^{-5}\).

For RL, we draw eight off-policy rollouts per example from the SFT checkpoint on TimeLens2-93K and TimeLens-100K, up-weighting examples with low mean tIoU. GRPO then uses eight generations per prompt, a global batch size of 64, a learning rate of \(10^{-6}\), and a KL coefficient of \(0.04\); group rewards are mean-centered without variance scaling. Videos are sampled at 2 fps, capped at 512 frames, and processed with a 16K-token budget. The default reward is \(R_{\mathrm{tIoU}} + R_{\mathrm{TW}} - \mathbf{1}_{\mathrm{invalid}}\); all RL experiments and ablations retain the fixed \(-\mathbf{1}_{\mathrm{invalid}}\) penalty for unparsable outputs. The visual encoder remains frozen in both stages, while the language model is updated.

\subsection{Comparison with State-of-the-Art}

\paragraph{Temporal grounding.}
\begin{table*}[t]
    \centering
    \colorlet{oursrowfaint}{reporthighlight!55!white}
    \colorlet{oursrowlight}{reporthighlight}
    \colorlet{oursrowdark}{reporthighlightstrong}
    \colorlet{baserowlight}{reportbasehighlight}
    \colorlet{baserowdark}{reportbasehighlightstrong}
    \colorlet{basecaptionblue}{reportaccentdark}
    \colorlet{tablegroupgreen}{reportaccentdark}
    \newcommand{\modelscaleseparator}{%
        \arrayrulecolor{reportline!80!white}%
        \cmidrule[0.22pt](lr){2-16}%
        \arrayrulecolor{reportline}%
    }
    \caption{Comparison with proprietary and open-source models on temporal video grounding benchmarks. Metrics are R1@0.5 and mIoU (\%). \textsuperscript{TL} denotes evaluation on TimeLens-Bench~\citep{zhang2026timelens}; \(\dagger\) marks the VUE-TR vision subsets, and \(\ddagger\) the MomentSeeker text-only query subset.}
    \label{tab:temporal-grounding-benchmarks}
    \scriptsize
    \setlength{\tabcolsep}{2pt}
    \renewcommand{\arraystretch}{1.00}
    \resizebox{\textwidth}{!}{%
    \begin{tabular}{@{}c@{\hspace{2.5pt}}l*{14}{c}@{}}
        \toprule
        \multirow{2}{*}{\textbf{Size}} &
        \multirow{2}{*}{\textbf{Model}} &
        \multicolumn{2}{c}{\textbf{Charades}\textsuperscript{TL}} &
        \multicolumn{2}{c}{\textbf{ActivityNet}\textsuperscript{TL}} &
        \multicolumn{2}{c}{\textbf{QVHighlights}\textsuperscript{TL}} &
        \multicolumn{2}{c}{\textbf{VUE-TR}$^{\dag}$} &
        \multicolumn{2}{c}{\textbf{VUE-TR-V2}$^{\dag}$} &
        \multicolumn{2}{c}{\textbf{MomentSeeker}$^{\ddagger}$} &
        \multicolumn{2}{c}{\textbf{Ego4D-NLQ}} \\
        \cmidrule(lr){3-4} \cmidrule(lr){5-6} \cmidrule(lr){7-8} \cmidrule(lr){9-10} \cmidrule(lr){11-12} \cmidrule(lr){13-14} \cmidrule(lr){15-16}
        & &
        R1@0.5 & mIoU &
        R1@0.5 & mIoU &
        R1@0.5 & mIoU &
        R1@0.5 & mIoU &
        R1@0.5 & mIoU &
        R1@0.5 & mIoU &
        R1@0.5 & mIoU \\
        \midrule
        \multicolumn{2}{@{}l}{\reporttablegroup{Proprietary models}} & \multicolumn{14}{c}{} \\
        & GPT-5~\citep{gpt5} & 42.0 & 40.5 & 44.9 & 42.9 & 60.4 & 56.8 & \reportna & \reportna & 19.5 & 20.0 & \reportna & \reportna & \reportna & \reportna \\
        & Gemini 3 Pro~\citep{gemini3} & \reportna & \reportna & \reportna & \reportna & \reportna & \reportna & \reportna & \reportna & 41.3 & 39.7 & \reportna & \reportna & \reportna & \reportna \\
        & Gemini 2.5 Pro~\citep{Gemini2.5} & 61.1 & 52.8 & 64.2 & 58.1 & 75.9 & 70.4 & 20.5 & 21.9 & \reportna & \reportna & \reportna & \reportna & \reportna & \reportna \\
        \midrule
        \multicolumn{2}{@{}l}{\reporttablegroup{Open-source models}} & \multicolumn{14}{c}{} \\
        \multirow{4}{*}{\rotatebox[origin=c]{90}{\sffamily\bfseries\color{reportslate}2B}} & Marlin-2B~\citep{nemostation2026marlin2b} & 51.7 & 46.5 & 40.5 & 37.9 & 48.5 & 46.8 & 23.6 & 24.8 & 19.9 & 19.2 & 15.2 & 16.3 & 7.8 & 8.9 \\
        & Qwen3.5-2B~\citep{qwen35blog} & 46.8 & 46.0 & 52.8 & 48.9 & 65.7 & 60.6 & 27.2 & 27.8 & 21.5 & 21.6 & 12.8 & 15.6 & 10.1 & 11.2\\
        & Qwen3-VL-2B~\citep{qwen3vl} & 44.4 & 43.4 & 41.0 & 39.9 & 53.4 & 52.2 & 29.9 & 31.4 & 22.5 & 24.1 & 9.8 & 12.4 & 8.1 & 9.3 \\
        \rowcolor{oursrowfaint}\cellcolor{white} & \textbf{\methodname-2B} & 60.6 & 53.4 & 62.0 & 55.4 & 72.8 & 67.0 & 50.7 & 51.1 & 44.5 & 43.8 & 21.9 & 24.3 & 16.5 & 16.8 \\
        \modelscaleseparator
        \multirow{5}{*}{\rotatebox[origin=c]{90}{\sffamily\bfseries\color{reportslate}4B}} & InternVL3.5-4B~\citep{internvl3.5} & 14.1 & 16.0 & 12.7 & 14.9 & 15.8 & 17.7 & 14.1 & 16.5 & 6.0 & 8.5 & 2.1 & 3.2 & 0.7 & 2.0 \\
        & Qwen3-VL-4B~\citep{qwen3vl} & 53.8 & 49.4 & 54.8 & 50.7 & 66.4 & 62.3 & 33.0 & 33.6 & 19.9 & 20.3 & 13.5 & 15.3 & 10.6 & 11.4 \\
        & Molmo2-4B~\citep{Molmo2} & 31.1 & 34.7 & 40.3 & 40.9 & 62.6 & 60.8 & 42.5 & 43.1 & 30.9 & 32.1 & 14.7 & 19.5 & 7.7 & 10.0 \\
        & VideoChat3-4B~\citep{VideoChat3} & 64.9 & 56.1 & 60.5 & 54.6 & 72.5 & 67.0 & 47.8 & 47.9 & 40.4 & 40.2 & 24.4 & 25.9 & 13.9 & 14.6 \\
        \rowcolor{oursrowlight}\cellcolor{white} & \textbf{\methodname-4B} & 66.6 & 57.7 & 65.5 & 59.0 & 75.2 & 69.3 & 52.4 & 53.2 & 49.5 & 48.1 & 27.0 & 27.9 & 18.1 & 18.6 \\
        \modelscaleseparator
        \multirow{14}{*}{\rotatebox[origin=c]{90}{\sffamily\bfseries\color{reportslate}7B+}} & Qwen3.5-397B-A17B~\cite{qwen35blog} & 47.8 & 47.5 & 59.1 & 53.8 & 71.6 & 65.8 & 42.7 & 42.2 & 35.5 & 34.5 & 22.3 & 23.3 & 13.3 & 14.5 \\
        & Qwen3-VL-235B-A22B~\citep{qwen3vl} & 50.8 & 47.8 & 57.5 & 52.2 & 70.2 & 64.6 & 42.5 & 43.1 & 34.5 & 34.1 & 21.2 & 23.7 & 11.5 & 12.7 \\
        & Qwen3.5-35B-A3B~\cite{qwen35blog} & 50.4 & 48.2 & 58.6& 52.6 & 72.6 & 66.0 & 46.3 & 44.6 & 37.1 & 35.3&  20.2 & 22.6 & 12.0 & 13.0 \\
        & Qwen3-VL-30B-A3B~\citep{qwen3vl} & 46.5 & 48.1 & 51.1 & 49.4 & 67.6 & 63.2 & 40.4 & 42.5 & 32.3 & 32.6 & 19.6 & 22.6 & 10.4 & 11.5\\
        & Vidi-1.5-9B~\citep{vidi} & 18.3 & 22.0 & 40.2 & 39.3 & 58.4 & 55.9 & 45.0 & 47.0 & 31.9 & 34.3 & 14.8 & 18.6 & 6.5 & 8.4 \\
        & LLaVA-OneVision-2-8B~\citep{LLaVA-OneVision-2} & 59.6 & 52.6 & 57.9 & 52.4 & 70.0 & 65.7 & 40.8 & 41.3 & 35.0 & 34.5 & 18.0 & 19.6 & 12.0 & 12.8 \\
        & Qwen3-VL-8B~\citep{qwen3vl} & 49.3 & 47.2 & 51.1 & 48.0 & 62.9 & 59.4 & 23.8 & 25.5 & 13.1 & 14.0 & 9.4 & 9.8 & 4.8 & 5.3 \\
        & InternVideo3-8B~\citep{InternVideo3} & 61.7 & 53.2 & 51.4 & 46.3 & 62.6 & 59.2 & 39.4 & 41.1 & 32.4 & 32.6 & 17.9 & 19.3 & 7.3 & 8.6 \\
        & TimeLens-8B~\citep{zhang2026timelens} & 65.8 & 57.0 & 62.0 & 56.3 & 71.7 & 66.8 & 42.9 & 43.6 & 34.4 & 34.0 & 20.4 & 21.9 & 14.6 & 15.7 \\
        & Video-o3-7B~\citep{Video-o3} & 34.2 & 38.6 & 41.0 & 38.9& 49.5 & 47.4 & 22.1 & 25.1 & 12.9 & 15.3 & 9.8 & 13.0 & 1.9 & 3.1 \\
        & VideoChat-Flash-7B~\citep{videochat-flash} & 37.9 & 39.7 & 21.8 & 24.8 & 30.6 & 32.7 & 15.4 & 17.3 & 9.5 & 10.1 & 5.9 & 7.2 & 1.5 & 2.5 \\
        & MiMo-VL-7B~\citep{MiMo-VL} & 42.6 & 39.6 & 38.7 & 35.5 & 42.6 & 41.5 & 17.5 & 19.1 & 10.1 & 10.4 & 3.8 & 6.0 & 0.4 & 1.0 \\
        & TimeSuite-7B~\citep{timesuite} & 35.5 & 38.1 & 17.5 & 19.8 & 16.9 & 21.7 & 10.3 & 13.2 & 5.3 & 7.3 & 3.3 & 5.9 & 0.6 & 1.4 \\
        \rowcolor{oursrowdark}\cellcolor{white} & \textbf{\methodname-8B} & 68.0 & 58.6 & 65.6 & 58.6 & 76.1 & 70.2 & 51.6 & 53.5 & 49.1 & 47.7 & 27.3 & 28.5 & 18.4 & 19.0 \\
        \bottomrule
    \end{tabular}%
    }
\end{table*}

\Cref{tab:temporal-grounding-benchmarks} shows that \methodname performs consistently across model scales and benchmark regimes. At 2B, \methodname-2B outperforms all three size-matched open-source baselines on all seven benchmarks in both metrics, reaching 44.5 average mIoU---14.2 points above Qwen3-VL-2B~\citep{qwen3vl} and 2.4 points above TimeLens-8B~\citep{zhang2026timelens}. \methodname-8B surpasses all prior methods in R1@0.5 on all seven benchmarks and in mIoU on six, trailing Gemini 2.5 Pro on QVHighlights by only 0.2 points. Notably, \methodname-4B already outperforms every prior model, including substantially larger and proprietary systems, on six of seven benchmarks in both metrics. Its advantage is largest where temporal search is hardest: over the Qwen3-VL-4B backbone, it improves mIoU by 19.6 points on VUE-TR and 27.8 on VUE-TR-V2, while gaining 12.6 on question-form MomentSeeker and 7.2 on egocentric Ego4D-NLQ. These gains across long videos, query forms, and viewpoints indicate that \methodname improves temporal evidence search rather than merely fitting dataset-specific boundary patterns.

\begin{figure}[t]
    \centering
    \includegraphics[width=\textwidth]{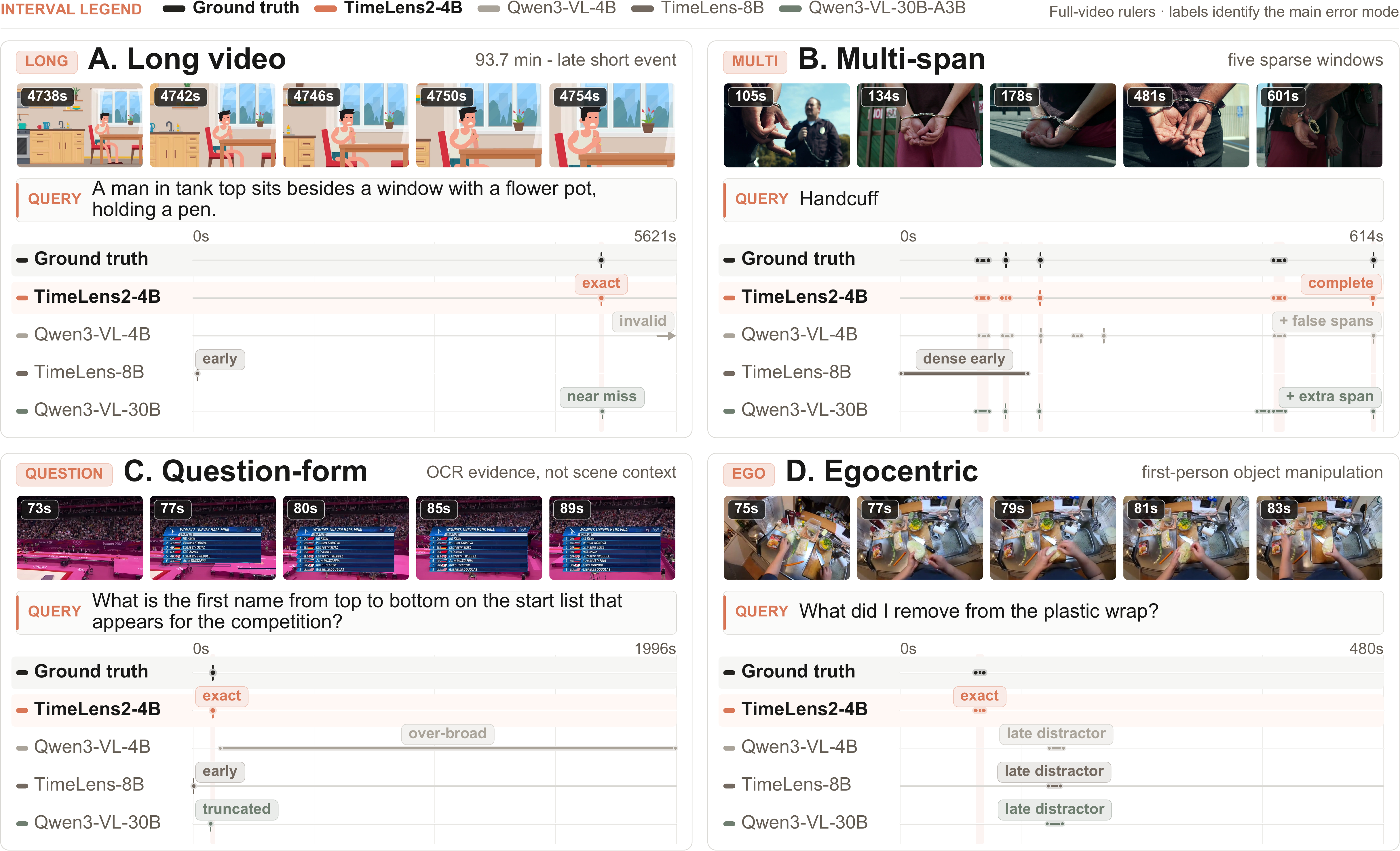}
    \caption{\textbf{Qualitative comparison on temporal grounding.} Each panel shows five target frames, the query, and a full-video ruler from \(0\) to the video duration; colored spans follow the shared legend. Dense labels summarize many short windows after merging small display gaps.}
    \label{fig:qualitative-analysis}
\end{figure}

\paragraph{Qualitative analysis.}
The central challenge in temporal grounding is not event recognition alone, but selecting query-relevant evidence from long, repetitive, or fragmented timelines. \Cref{fig:qualitative-analysis} makes this distinction visible. In a 93.7-minute video, \methodname retrieves a brief event near 4,750 seconds, while the baselines drift, miss the target, or produce an invalid timestamp. When the evidence recurs, it recovers all five sparse handcuff spans without adding plausible but false windows. The question-form example further separates semantic relevance from answer-bearing evidence: \methodname isolates the brief OCR start list needed to answer the query rather than grounding the broader competition scene. In the egocentric case, it identifies the actual object removal while every baseline selects a later, visually similar interaction. Together, these cases show that \methodname learns evidence-directed temporal search: it remains complete across disjoint spans while resisting distractors, even when the evidence is late, sparse, or defined by the query's intent.

\subsection{Ablation Study}

\paragraph{Effect of TimeLens2-93K.}
We ask whether scaling TimeLens2-93K teaches temporal search itself or primarily broadens its coverage. To isolate supervision from RL, we fine-tune Qwen3-VL-4B on different data mixtures. \Cref{tab:timelens2-data-ablation} reveals a two-phase scaling curve. The first 5\% of TimeLens2-93K delivers the largest gain, raising average mIoU from 34.7 to 42.8; 20\% reaches 45.3. The full corpus improves the average more modestly to 45.8, but its gains concentrate on harder settings such as VUE-TR-V2 and Ego4D-NLQ. Thus, a small high-confidence subset establishes the basic evidence-search behavior, while broader coverage improves robustness to long contexts and ambiguous boundaries. This behavior also transfers across query forms: although TimeLens2-93K contains only declarative queries, it raises question-form MomentSeeker from 15.3 to 25.8 mIoU, indicating that the model learns to search for evidence rather than imitate a query template.

\begin{center}
\begin{minipage}{\textwidth}
    \centering
    \colorlet{dataablationhighlight}{reporthighlight}
    \colorlet{dataablationcore}{reportbasehighlight}
    \colorlet{tablegroupgreen}{reportaccentdark}
    \captionsetup{type=table,hypcap=false}
    \captionof{table}{\textbf{Effect of TimeLens2-93K in supervised training.} Qwen3-VL-4B mIoU (\%) under different data recipes; Avg. is unweighted. Sizes are videos / QA turns in thousands, with QA turns counted from multi-turn annotations. Shading marks the core corpus and final mixture.}
    \vspace{-0.8em}
    \label{tab:timelens2-data-ablation}
    \scriptsize
    \setlength{\tabcolsep}{3.0pt}
    \renewcommand{\arraystretch}{1.11}
    \resizebox{\textwidth}{!}{%
    \begin{tabular}{@{}lc@{\hspace{0.75em}}cccccccc@{}}
        \toprule
        \multirow{2}{*}{\raisebox{-1ex}{\textbf{Supervised setting}}} &
        \multirow{2}{*}{\raisebox{-4ex}{\shortstack{\textbf{Videos / QA}\\\textbf{(K)}}}} &
        \multicolumn{8}{c}{\textbf{Evaluation mIoU (\%)}} \\
        \cmidrule(lr){3-10}
        &
        &
        \textbf{Charades}\textsuperscript{TL} &
        \textbf{ActivityNet}\textsuperscript{TL} &
        \textbf{QVHighlights}\textsuperscript{TL} &
        \textbf{VUE-TR}$^{\dag}$ &
        \textbf{VUE-TR-V2}$^{\dag}$ &
        \textbf{MomentSeeker}$^{\ddagger}$ &
        \textbf{Ego4D-NLQ} &
        \textbf{Avg.} \\
        \midrule
        \multicolumn{10}{@{}l}{\reporttablegroup{Backbone}} \\
        \hspace{0.6em}Qwen3-VL-4B & \reportna & 49.4 & 50.7 & 62.3 & 33.6 & 20.3 & 15.3 & 11.4 & 34.7 \\
        \addlinespace[-0.15em]
        \midrule
        \multicolumn{10}{@{}l}{\reporttablegroup{Scaling TimeLens2-93K}} \\
        \hspace{0.6em}5\% TimeLens2-93K & 1.2 / 4.7 & 53.0 & 55.7 & 67.5 & 48.3 & 38.0 & 22.8 & 14.2 & 42.8 \\
        \hspace{0.6em}20\% TimeLens2-93K & 4.8 / 18.6 & 53.4 & 57.2 & 68.8 & \textbf{51.1} & 44.4 & \textbf{26.2} & 15.8 & 45.3 \\
        \hspace{0.6em}50\% TimeLens2-93K & 11.9 / 46.6 & 53.4 & 57.4 & 68.6 & 50.5 & 45.2 & 26.1 & 15.9 & 45.3 \\
        \rowcolor{dataablationcore}\hspace{0.6em}\textbf{100\% TimeLens2-93K} & 23.8 / 93.2 & 54.7 & 57.7 & 68.8 & 50.3 & \textbf{46.0} & 25.8 & 17.2 & 45.8 \\
        \addlinespace[-0.15em]
        \midrule
        \multicolumn{10}{@{}l}{\reporttablegroup{Dataset comparison and mixture}} \\
        \addlinespace[0.08em]
        \hspace{0.6em}TimeLens-100K & 19.4 / 96.6 & 53.0 & 51.6 & 61.3 & 39.0 & 33.8 & 22.9 & 14.5 & 39.4 \\
        \addlinespace[0.05em]
        \shortstack[l]{\hspace{0.6em}TimeLens2-93K\\\hspace{0.6em}~+ TimeLens-100K} & \raisebox{1.15ex}{43.2 / 189.8} & \raisebox{1.15ex}{55.9} & \raisebox{1.15ex}{58.4} & \raisebox{1.15ex}{69.0} & \raisebox{1.15ex}{50.3} & \raisebox{1.15ex}{45.3} & \raisebox{1.15ex}{25.8} & \raisebox{1.15ex}{17.3} & \raisebox{1.15ex}{46.0} \\
        \addlinespace[0.05em]
        \rowcolor{dataablationhighlight}\shortstack[l]{\hspace{0.6em}\textbf{TimeLens2-93K}\\[-0.2ex]\hspace{0.6em}~\textbf{+ TimeLens-100K}\\[-0.2ex]\hspace{0.6em}~\textbf{+ Ego4D-NLQ}} & \raisebox{2.0ex}{\textbf{44.5 / 203.7}} & \raisebox{2.0ex}{\textbf{56.3}} & \raisebox{2.0ex}{\textbf{58.9}} & \raisebox{2.0ex}{\textbf{69.2}} & \raisebox{2.0ex}{50.9} & \raisebox{2.0ex}{45.6} & \raisebox{2.0ex}{25.8} & \raisebox{2.0ex}{\textbf{17.8}} & \raisebox{2.0ex}{\textbf{46.4}} \\
        \bottomrule
    \end{tabular}%
    }
\end{minipage}
\end{center}

Scale alone does not explain the gain: TimeLens-100K has a comparable number of QA turns but reaches only 39.4 average mIoU, with substantially weaker long-video and QVHighlights results. Adding it to TimeLens2-93K yields a smaller improvement from 45.8 to 46.0, mainly on the TimeLens-Bench-style datasets; Ego4D-NLQ then raises the average to 46.4 and strengthens egocentric grounding. TimeLens2-93K therefore provides the transferable search capability, while TimeLens-100K and Ego4D-NLQ add complementary benchmark and viewpoint coverage.

\paragraph{Impact of progressive label curation.}
Effective curation should remove distinct supervision errors, not merely shrink the dataset. Under a fixed Qwen3-VL-4B SFT recipe, \Cref{tab:label-curation-ablation} reveals a quality-over-quantity pattern. The raw annotators tie at 42.0 average mIoU yet differ by dataset, making agreement informative. Temporal consensus cuts the Qwen-anchored pool from 735.4K to 174.2K QA turns while raising mIoU to 43.4; semantic verification leaves 93.2K and reaches 44.1. Removing temporal instability and semantic mismatch therefore improves performance with only one-eighth as many labels.

With evidence identity established, boundary noise becomes the dominant bottleneck. Refining the same 93.2K labels yields the largest stage gain, adding 1.7 points to reach 45.8 mIoU without additional data. The full cascade improves over the raw Qwen labels by 3.8 points overall, including 6.8 on ActivityNet and 8.7 on QVHighlights. Although refinement loses 0.5 points on VUE-TR and MomentSeeker, the progression exposes a clear factorization of annotation error: consensus tests temporal reproducibility, semantic verification tests relevance, and local refinement resolves boundary uncertainty.

\begin{center}
\begin{minipage}{\textwidth}
    \centering
    \colorlet{labelcurationhighlight}{reporthighlight}
    \colorlet{labelanchorhighlight}{reportbasehighlight}
    \colorlet{labelgaincolor}{reportpositive}
    \colorlet{labellosscolor}{reportnegative}
    \colorlet{labelsamecolor}{reportslate}
    \newcommand{\labelgain}[1]{{\scriptsize\textcolor{labelgaincolor}{(#1)}}}
    \newcommand{\labelloss}[1]{{\scriptsize\textcolor{labellosscolor}{(#1)}}}
    \newcommand{\labelsame}[1]{{\scriptsize\textcolor{labelsamecolor}{(#1)}}}
    \captionsetup{type=table,hypcap=false}
    \captionof{table}{\textbf{Impact of progressive label curation.} Qwen3-VL-4B mIoU (\%) under a fixed SFT recipe. Parentheses report stage-wise changes in the Qwen-anchored cascade. The cascade applies temporal consensus, semantic verification, and boundary refinement. Shading marks the Qwen anchor and final labels.}
    \vspace{-0.8em}
    \label{tab:label-curation-ablation}
    \footnotesize
    \setlength{\tabcolsep}{2.8pt}
    \renewcommand{\arraystretch}{1.08}
    \resizebox{\textwidth}{!}{%
    \begin{tabular}{@{}lc@{\hspace{0.75em}}cccccccc@{}}
        \toprule
        \multirow{2}{*}{\raisebox{-1ex}{\textbf{Label curation stage}}} &
        \multirow{2}{*}{\raisebox{-4ex}{\shortstack{\textbf{Videos / QA}\\\textbf{(K)}}}} &
        \multicolumn{8}{c}{\textbf{Evaluation mIoU (\%)}} \\
        \cmidrule(lr){3-10}
        &
        &
        \textbf{Charades}\textsuperscript{TL} &
        \textbf{ActivityNet}\textsuperscript{TL} &
        \textbf{QVHighlights}\textsuperscript{TL} &
        \textbf{VUE-TR}$^{\dag}$ &
        \textbf{VUE-TR-V2}$^{\dag}$ &
        \textbf{MomentSeeker}$^{\ddagger}$ &
        \textbf{Ego4D-NLQ} &
        \textbf{Avg.} \\
        \midrule
        \rowcolor{labelanchorhighlight}\textbf{Raw Qwen3-VL-30B-A3B labels} & 28.9 / 735.4 & 51.7 & 50.9 & 60.1 & 47.4 & 43.2 & 24.9 & 15.7 & 42.0 \\
        Raw TimeLens-8B labels & 33.6 / 999.2 & 54.4 & 51.5 & 59.2 & 48.5 & 40.3 & 24.7 & 15.7 & 42.0 \\
        \midrule
        \multicolumn{10}{@{}l}{\reporttablegroup{Qwen-anchored curation cascade}} \\
        \hspace{0.6em}+ Temporal consensus & 26.4 / 174.2 & 53.1\,\labelgain{+1.4} & 52.9\,\labelgain{+2.0} & 62.0\,\labelgain{+1.9} & 49.5\,\labelgain{+2.1} & 44.1\,\labelgain{+0.9} & 25.8\,\labelgain{+0.9} & 16.6\,\labelgain{+0.9} & 43.4\,\labelgain{+1.4} \\
        \hspace{0.6em}+ Semantic verification & 23.8 / 93.2 & 53.3\,\labelgain{+0.2} & 53.4\,\labelgain{+0.5} & 63.0\,\labelgain{+1.0} & 50.8\,\labelgain{+1.3} & 45.1\,\labelgain{+1.0} & 26.3\,\labelgain{+0.5} & 16.6\,\labelsame{+0.0} & 44.1\,\labelgain{+0.6} \\
        \rowcolor{labelcurationhighlight}\hspace{0.6em}\textbf{+ Boundary refinement} & \textbf{23.8 / 93.2} & \textbf{54.7}\,\labelgain{+1.4} & \textbf{57.7}\,\labelgain{+4.3} & \textbf{68.8}\,\labelgain{+5.8} & 50.3\,\labelloss{-0.5} & \textbf{46.0}\,\labelgain{+0.9} & 25.8\,\labelloss{-0.5} & \textbf{17.2}\,\labelgain{+0.6} & \textbf{45.8}\,\labelgain{+1.7} \\
        \bottomrule
    \end{tabular}%
    }
\end{minipage}
\end{center}

\paragraph{Effect of long-context training.}
Longer contexts help only if the model can search them without absorbing more distractors. Holding the data, backbone, and optimization fixed, we vary the maximum packed length. \Cref{tab:long-context-ablation} shows average mIoU rising from 44.4 at 16K to 46.4 at 100K, but unevenly: Charades and QVHighlights peak by 32K, whereas VUE-TR-V2, MomentSeeker, and Ego4D-NLQ gain 4.0, 1.5, and 3.3 points by 100K. VUE-TR's late 1.8-point jump from 64K to 100K further suggests a context-threshold effect for long-range examples. These asymmetries indicate that long-context training expands the effective temporal search horizon rather than uniformly benefiting every task. We therefore use 100K for final SFT.

\begin{center}
\begin{minipage}{\textwidth}
    \centering
    \colorlet{contextablationhighlight}{reporthighlight}
    \colorlet{contextgaincolor}{reportpositive}
    \colorlet{contextlosscolor}{reportnegative}
    \colorlet{contextsamecolor}{reportslate}
    \newcommand{\contextgain}[1]{{\tiny\textcolor{contextgaincolor}{(#1)}}}
    \newcommand{\contextloss}[1]{{\tiny\textcolor{contextlosscolor}{(#1)}}}
    \newcommand{\contextsame}[1]{{\tiny\textcolor{contextsamecolor}{(#1)}}}
    \captionsetup{type=table,hypcap=false}
    \captionof{table}{\textbf{Effect of long-context training.} Qwen3-VL-4B mIoU (\%) with fixed supervised data and varying maximum packed length. Shading marks the final recipe.}
    \vspace{-0.8em}
    \label{tab:long-context-ablation}
    \scriptsize
    \setlength{\tabcolsep}{3.2pt}
    \renewcommand{\arraystretch}{1.08}
    \resizebox{\textwidth}{!}{%
    \begin{tabular}{lcccccccc}
        \toprule
        \textbf{Max. packed length} &
        \textbf{Charades}\textsuperscript{TL} &
        \textbf{ActivityNet}\textsuperscript{TL} &
        \textbf{QVHighlights}\textsuperscript{TL} &
        \textbf{VUE-TR}$^{\dag}$ &
        \textbf{VUE-TR-V2}$^{\dag}$ &
        \textbf{MomentSeeker}$^{\ddagger}$ &
        \textbf{Ego4D-NLQ} &
        \textbf{Avg.} \\
        \midrule
        16K & 56.1 & 57.0 & 68.2 & 49.1 & 41.6 & 24.3 & 14.5 & 44.4 \\
        32K & \textbf{56.3}\,\contextgain{+0.2} & 57.5\,\contextgain{+0.5} & \textbf{69.2}\,\contextgain{+1.0} & 48.8\,\contextloss{-0.3} & 44.2\,\contextgain{+2.6} & 25.1\,\contextgain{+0.8} & 16.6\,\contextgain{+2.1} & 45.4\,\contextgain{+1.0} \\
        64K & 56.1\,\contextloss{-0.2} & 58.5\,\contextgain{+1.0} & \textbf{69.2}\,\contextsame{+0.0} & 49.1\,\contextgain{+0.3} & 45.5\,\contextgain{+1.3} & 25.7\,\contextgain{+0.6} & 17.4\,\contextgain{+0.8} & 45.9\,\contextgain{+0.5} \\
        \rowcolor{contextablationhighlight}100K & \textbf{56.3}\,\contextgain{+0.2} & \textbf{58.9}\,\contextgain{+0.4} & \textbf{69.2}\,\contextsame{+0.0} & \textbf{50.9}\,\contextgain{+1.8} & \textbf{45.6}\,\contextgain{+0.1} & \textbf{25.8}\,\contextgain{+0.1} & \textbf{17.8}\,\contextgain{+0.4} & \textbf{46.4}\,\contextgain{+0.5} \\
        \bottomrule
    \end{tabular}%
    }
\end{minipage}
\end{center}

\paragraph{Effect of instruction and response-format diversity.}
A generalist grounder should recover the same evidence regardless of request phrasing or timestamp serialization; otherwise, it may learn the interface rather than the task. Holding the TimeLens2-93K examples, Qwen3-VL-4B backbone, and SFT recipe fixed, we independently vary requests and renderings using the 27 instructions and 28 response specifications in \Cref{tab:instruction-response-format-diversity}.

\Cref{tab:instruction-response-format-ablation} shows that rendering diversity alone raises average mIoU from 44.9 to 45.4, while request diversity reaches 45.0. Together they reach 45.8 (+0.9), perform best on six of seven benchmarks, and yield the largest gains on VUE-TR (+3.0) and QVHighlights (+1.3). The joint gain exceeds the sum of the isolated gains by 0.3 points. Thus, response-format diversity is the stronger standalone lever, but their positive interaction suggests that varying both sides of the interface helps disentangle evidence localization from surface realization and improves transfer across prompting protocols.

\begin{center}
\begin{minipage}{\textwidth}
    \centering
    \colorlet{formatablationhighlight}{reporthighlight}
    \colorlet{formatbaselinehighlight}{reportbasehighlight}
    \colorlet{formatgaincolor}{reportpositive}
    \newcommand{\formatgain}[1]{{\tiny\textcolor{formatgaincolor}{(#1)}}}
    \captionsetup{type=table,hypcap=false}
    \captionof{table}{\textbf{Effect of instruction and response-format diversity.} Qwen3-VL-4B mIoU (\%) in a $2\!\times\!2$ ablation with fixed examples and SFT recipe. ``Single'' fixes the request or numeric interval-array rendering; ``Diverse'' samples from the full pools. Parentheses in Avg. show gains over the fully fixed baseline; shading marks the baseline and fully diverse setting.}
    \vspace{-0.8em}
    \label{tab:instruction-response-format-ablation}
    \scriptsize
    \setlength{\tabcolsep}{2.8pt}
    \renewcommand{\arraystretch}{1.10}
    \resizebox{\textwidth}{!}{%
    \begin{tabular}{@{}llcccccccc@{}}
        \toprule
        \textbf{Grounding request} &
        \textbf{Target rendering} &
        \textbf{Charades}\textsuperscript{TL} &
        \textbf{ActivityNet}\textsuperscript{TL} &
        \textbf{QVHighlights}\textsuperscript{TL} &
        \textbf{VUE-TR}$^{\dag}$ &
        \textbf{VUE-TR-V2}$^{\dag}$ &
        \textbf{MomentSeeker}$^{\ddagger}$ &
        \textbf{Ego4D-NLQ} &
        \textbf{Avg.} \\
        \midrule
        \rowcolor{formatbaselinehighlight}\textbf{Single} & \textbf{Single} & 54.3 & 57.0 & 67.5 & 47.3 & 45.8 & 25.7 & 16.6 & 44.9 \\
        Single & Diverse & \textbf{54.8} & 57.4 & 68.5 & 50.0 & 45.4 & 25.3 & 16.3 & 45.4\,\formatgain{+0.5} \\
        Diverse & Single & 54.7 & 57.2 & 67.3 & 48.2 & 45.7 & 25.3 & 16.5 & 45.0\,\formatgain{+0.1} \\
        \rowcolor{formatablationhighlight}\textbf{Diverse} & \textbf{Diverse} & 54.7 & \textbf{57.7} & \textbf{68.8} & \textbf{50.3} & \textbf{46.0} & \textbf{25.8} & \textbf{17.2} & \textbf{45.8}\,\formatgain{+0.9} \\
        \bottomrule
    \end{tabular}%
    }
\end{minipage}
\end{center}

\paragraph{Ablation on temporal reward design.}
\begin{wrapfigure}{r}{0.43\linewidth}
    \centering
    \vspace{-1.2em}
    \includegraphics[width=\linewidth]{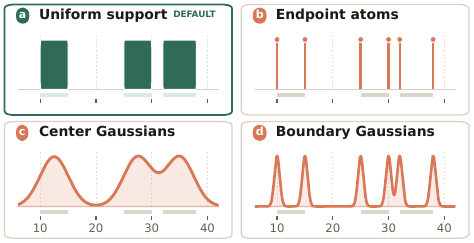}
    \vspace{-1.9em}
    \caption{\textbf{Temporal mass models.} The same intervals \([[10,15], [25,30], [32,38]]\) are converted into four probability distributions for the Wasserstein reward.}
    \label{fig:temporal-reward-distributions}
    \vspace{-0.8em}
\end{wrapfigure}
\(R_{\mathrm{TW}}\) is only as informative as the temporal distribution used to represent an interval set. We therefore ask whether the reward should preserve full interval occupancy or compress it into endpoints, centers, or boundaries. Holding all other RL settings fixed, including the parse penalty, we compare the four mass models in \Cref{fig:temporal-reward-distributions}. As shown in \Cref{tab:temporal-reward-ablation}, every variant improves average mIoU over \(R_{\mathrm{tIoU}}\), from 47.0 to 47.4--47.7, confirming that distance-aware credit robustly complements overlap. Uniform support is the only variant that improves every benchmark and achieves the best average, 47.7.

\begin{center}
\begin{minipage}{\textwidth}
    \centering
    \colorlet{rewardbaselinehighlight}{reportbasehighlight}
    \colorlet{rewardgaincolor}{reportpositive}
    \newcommand{\rewardgain}[1]{{\tiny\textcolor{rewardgaincolor}{(#1)}}}
    \captionsetup{type=table,hypcap=false}
    \captionof{table}{\textbf{Ablation on temporal reward design.} mIoU (\%) when varying only \(R_{\mathrm{temporal}}\) in \(R=R_{\mathrm{temporal}}-\mathbf{1}_{\mathrm{invalid}}\); all other RL settings are fixed. Parentheses in Avg. show gains over the tIoU baseline; shading marks the baseline and default uniform-support design.}
    \vspace{-0.8em}
    \label{tab:temporal-reward-ablation}
    \scriptsize
    \setlength{\tabcolsep}{3.2pt}
    \renewcommand{\arraystretch}{1.08}
    \resizebox{\textwidth}{!}{%
    \begin{tabular}{llcccccccc}
        \toprule
        \textbf{Temporal reward} & \textbf{Temporal distribution} &
        \textbf{Charades}\textsuperscript{TL} &
        \textbf{ActivityNet}\textsuperscript{TL} &
        \textbf{QVHighlights}\textsuperscript{TL} &
        \textbf{VUE-TR}$^{\dag}$ &
        \textbf{VUE-TR-V2}$^{\dag}$ &
        \textbf{MomentSeeker}$^{\ddagger}$ &
        \textbf{Ego4D-NLQ} &
        \textbf{Avg.} \\
        \midrule
        \rowcolor{rewardbaselinehighlight}\(R_{\mathrm{tIoU}}\) & \reportna & 57.4 & 58.1 & 67.2 & 52.0 & 47.8 & 27.7 & 18.5 & 47.0 \\
        \rowcolor{reporthighlight}\(R_{\mathrm{tIoU}} + R_{\mathrm{TW}}\) & \textbf{Uniform support} & \textbf{57.7} & \textbf{59.0} & 69.3 & 53.2 & \textbf{48.1} & 27.9 & 18.6 & \textbf{47.7}\,\rewardgain{+0.7} \\
        \(R_{\mathrm{tIoU}} + R_{\mathrm{TW}}\) & Endpoint atoms & 57.2 & 58.9 & 68.4 & \textbf{53.9} & 47.6 & 27.6 & 18.5 & 47.4\,\rewardgain{+0.4} \\
        \(R_{\mathrm{tIoU}} + R_{\mathrm{TW}}\) & Center Gaussians & 57.4 & \textbf{59.0} & \textbf{69.4} & 53.2 & 47.4 & \textbf{28.3} & 18.8 & 47.6\,\rewardgain{+0.6} \\
        \(R_{\mathrm{tIoU}} + R_{\mathrm{TW}}\) & Boundary Gaussians & 57.1 & \textbf{59.0} & 69.2 & 52.9 & 46.9 & 28.0 & \textbf{18.9} & 47.4\,\rewardgain{+0.4} \\
        \bottomrule
    \end{tabular}%
    }
\end{minipage}
\end{center}

Uniform support is more consistent because it preserves both where the evidence lies and how long it lasts, matching the span occupancy measured by mIoU. Endpoint atoms and boundary Gaussians emphasize transitions, while center Gaussians mainly encode location; each discards part of the interval interior or duration. Such summaries may favor individual datasets but are less stable across tasks. The broader lesson is that dense credit alone is insufficient: the reward representation should preserve the geometry of the evaluation object. We therefore use uniform support by default.

\paragraph{Comparison with matched NGIoU.}
NGIoU also provides graded feedback for non-overlapping intervals, making it a natural alternative to \(R_{\mathrm{TW}}\). We replace only \(R_{\mathrm{TW}}\) with \(R_{\mathrm{NGIoU}}\), using MUSEG's one-to-one matching~\citep{MUSEG}. As shown in \Cref{tab:temporal-reward-ngiou-ablation}, NGIoU reaches 47.1 average mIoU, compared with 47.7 for temporal Wasserstein. The gap is larger on the multi-interval VUE-TR benchmarks: 1.4 points on VUE-TR and 1.0 on VUE-TR-V2. NGIoU depends on pairwise assignments, which can change when predictions split, merge, or differ in number. In contrast, \(R_{\mathrm{TW}}\) compares merged temporal support directly and is invariant to equivalent fragmentations. Thus, dense feedback alone is insufficient; robust multi-interval rewards should avoid fragile correspondences.

\begin{center}
\begin{minipage}{\textwidth}
    \centering
    \colorlet{ngiouablationhighlight}{reporthighlight}
    \colorlet{ngioucomparehighlight}{reportbasehighlight}
    \colorlet{ngiougaincolor}{reportpositive}
    \newcommand{\ngiougain}[1]{{\tiny\textcolor{ngiougaincolor}{(#1)}}}
    \captionsetup{type=table,hypcap=false}
    \captionof{table}{\textbf{Temporal Wasserstein vs. matched NGIoU.} We vary only the auxiliary dense reward; all other RL settings are fixed. \(R_{\mathrm{NGIoU}}\) uses MUSEG's start-time-sorted one-to-one matching~\citep{MUSEG}. The Avg. gain is relative to matched NGIoU; shading distinguishes the comparator and our default reward.}
    \vspace{-0.8em}
    \label{tab:temporal-reward-ngiou-ablation}
    \scriptsize
    \setlength{\tabcolsep}{3.2pt}
    \renewcommand{\arraystretch}{1.08}
    \resizebox{\textwidth}{!}{%
    \begin{tabular}{lcccccccc}
        \toprule
        \textbf{Temporal reward} &
        \textbf{Charades}\textsuperscript{TL} &
        \textbf{ActivityNet}\textsuperscript{TL} &
        \textbf{QVHighlights}\textsuperscript{TL} &
        \textbf{VUE-TR}$^{\dag}$ &
        \textbf{VUE-TR-V2}$^{\dag}$ &
        \textbf{MomentSeeker}$^{\ddagger}$ &
        \textbf{Ego4D-NLQ} &
        \textbf{Avg.} \\
        \midrule
        \rowcolor{ngiouablationhighlight}\(R_{\mathrm{tIoU}} + R_{\mathrm{TW}}\) & \textbf{57.7} & \textbf{59.0} & \textbf{69.3} & \textbf{53.2} & \textbf{48.1} & 27.9 & \textbf{18.6} & \textbf{47.7}\,\ngiougain{+0.6} \\
        \rowcolor{ngioucomparehighlight}\(R_{\mathrm{tIoU}} + R_{\mathrm{NGIoU}}\) & \textbf{57.7} & 58.8 & 67.7 & 51.8 & 47.1 & \textbf{28.2} & 18.5 & 47.1 \\
        \bottomrule
    \end{tabular}%
    }
\end{minipage}
\end{center}

\paragraph{Diagnostic ablation on zero-overlap misses.}
Average gains establish efficacy, but not mechanism: \(R_{\mathrm{TW}}\) may help for reasons unrelated to its distance-aware signal. We therefore examine the 4,332 benchmark predictions from RL-tIoU that have no overlap with the target and group them by their minimum temporal distance to it. If \(R_{\mathrm{TW}}\) supplies the intended geometry, recovery should decay with distance.

\begin{center}
\begin{minipage}{1\linewidth}
    \centering
    \colorlet{diagnostichighlight}{reporthighlight}
    \colorlet{diagnosticmidhighlight}{reporthighlight!50!white}
    \captionsetup{type=table,hypcap=false}
    \captionof{table}{\textbf{Where does temporal Wasserstein help?} We stratify the 4,332 valid zero-tIoU predictions from RL-tIoU by distance to the target. Gap is normalized by total ground-truth duration; recovery is the fraction of corresponding examples on which adding \(R_{\mathrm{TW}}\) yields positive tIoU. Shading follows the distance gradient.}
    \vspace{-0.5em}
    \label{tab:temporal-reward-zero-overlap}
    \small
    \setlength{\tabcolsep}{10pt}
    \renewcommand{\arraystretch}{1.12}
    \resizebox{0.55\linewidth}{!}{%
    \begin{tabular}{@{}lcccc@{}}
        \toprule
        \multicolumn{3}{c}{\textbf{Zero-overlap misses under \(R_{\mathrm{tIoU}}\)}} &
        \multicolumn{2}{c}{\textbf{After adding \(R_{\mathrm{TW}}\)}} \\
        \cmidrule(lr){1-3}\cmidrule(l){4-5}
        \textbf{Distance} &
        \textbf{Normalized gap} &
        \textbf{Cases} &
        \textbf{Recovered (\%)} $\uparrow$ &
        \textbf{mIoU (\%)} $\uparrow$ \\
        \midrule
        \rowcolor{diagnostichighlight}
        Near & $\leq 1\times$ & 644 & \textbf{21.9} & \textbf{7.5} \\
        \rowcolor{diagnosticmidhighlight}
        Mid & $1$--$5\times$ & 808 & 15.8 & 4.0 \\
        Far & $>5\times$ & 2,880 & 5.7 & 1.4 \\
        \midrule
        \multicolumn{2}{@{}l}{\textbf{All zero-overlap misses}} & \textbf{4,332} & \textbf{10.0} & \textbf{2.8} \\
        \bottomrule
    \end{tabular}%
    }
\end{minipage}
\end{center}

\Cref{tab:temporal-reward-zero-overlap} shows a clear distance effect. Adding \(R_{\mathrm{TW}}\) recovers positive overlap for 21.9\% of near misses, 15.8\% of mid-distance misses, and only 5.7\% of far misses; mIoU similarly falls from 7.5 to 4.0 and 1.4. These zero-overlap cases also account for most of the overall improvement from 47.0 to 47.7 in \Cref{tab:temporal-reward-ablation}. Thus, \(R_{\mathrm{TW}}\) helps mainly where tIoU provides no signal, giving more credit to near misses than distant errors.

\FloatBarrier
\paragraph{Diagnostic on group-relative reward structure.}
GRPO learns from within-group preferences, so a dense reward helps only if it breaks ties among rollouts. With tIoU, a group of non-overlapping predictions can receive uniformly zero reward and, after mean-centering, no temporal learning signal. To isolate the effect of \(R_{\mathrm{TW}}\), we score the same valid rollout groups collected over 300 training steps with both \(R_{\mathrm{tIoU}}\) and \(R_{\mathrm{tIoU}}+R_{\mathrm{TW}}\).

\begin{table}[htbp]
    \centering
    \colorlet{grpobaselinehighlight}{reportbasehighlight}
    \caption{\textbf{Does temporal Wasserstein create group-relative signal?} We score the same valid rollout groups collected over 300 training steps with both rewards. Constant groups contain only one reward value; zero-tIoU rescue is the fraction of all-zero-tIoU groups that become non-constant after adding \(R_{\mathrm{TW}}\). Advantages are mean-centered without variance scaling.}
    \vspace{-0.6em}
    \label{tab:grpo-group-variance}
    \scriptsize
    \setlength{\tabcolsep}{4.2pt}
    \renewcommand{\arraystretch}{1.08}
    \resizebox{0.92\textwidth}{!}{%
    \begin{tabular}{lccccc}
        \toprule
        \textbf{Scoring reward} &
        \textbf{Constant groups} \(\downarrow\) &
        \textbf{Distinct/group} \(\uparrow\) &
        \(\operatorname{Var}_{g}(R)\) \(\uparrow\) &
        \(\mathbb{E}_{g}|A|\) \(\uparrow\) &
        \textbf{Zero-tIoU rescue} \(\uparrow\) \\
        \midrule
        \rowcolor{grpobaselinehighlight}\(R_{\mathrm{tIoU}}\) & 13.8\% & 4.63 & 0.023 & 0.098 & \reportna \\
        \rowcolor{reporthighlight}\(R_{\mathrm{tIoU}}+R_{\mathrm{TW}}\) & \textbf{3.6\%} & \textbf{6.04} & \textbf{0.091} & \textbf{0.190} & \textbf{75.8\%} \\
        \bottomrule
    \end{tabular}%
    }
\end{table}

\FloatBarrier

As shown in \Cref{tab:grpo-group-variance}, adding \(R_{\mathrm{TW}}\) reduces constant-reward groups from 13.8\% to 3.6\% and rescues 75.8\% of all-zero-tIoU groups. Beyond tie-breaking, distinct rewards per group rise from 4.63 to 6.04, within-group variance increases by \(4.0\times\), and mean absolute advantage nearly doubles. Because rewards are mean-centered without variance scaling, this extra spread passes directly into the policy update. Temporal Wasserstein thus turns otherwise silent groups into ranked training signals, explaining its strong fit with group-relative optimization.

\section{Conclusion}
\label{sec:conclusion}

Video MLLMs become reliable interfaces to video archives only when their answers are traceable to supporting moments. This requires trustworthy evidence labels and an objective that guides imperfect interval predictions. \methodname addresses both. TimeLens2-93K replaces brittle global annotation with evidence proposal, independent localization, consensus and semantic verification, and boundary refinement; the temporal Wasserstein reward supplies dense, matching-free geometry when overlap is uninformative. Across seven benchmarks, this design enables compact 2B, 4B, and 8B models to outperform much larger open-source models in long-video, multi-span, question-form, and egocentric settings. More broadly, treating evidence consistently as an interval set from supervision through optimization turns temporal grounding from a specialized retrieval task into a native, auditable capability of generalist video MLLMs.

\newpage
\bibliography{iclr2026_conference}
\bibliographystyle{iclr2026_conference}

\appendix
\clearpage
\section{Supplementary Material}
\label{sec:supplementary}

\etocsetnexttocdepth{subsection}
\etocsettocstyle{\subsection*{Contents}}{}
\localtableofcontents

\subsection{Instruction and Response-Format Diversity}
\label{sec:supp-instruction-response-diversity}

Temporal grounding requires both locating relevant evidence and expressing intervals through a specified interface. Training with a single prompt--response convention can entangle localization with surface form, reducing robustness when the request phrasing or output format changes.

For each QA turn, we keep the video, synthesized query \(q\), and target interval set \(\mathcal{Y}\) fixed, then independently and uniformly sample one of 27 grounding requests and one of 28 response specifications. The request pool varies how localization is posed, while the response pool controls answer syntax, timestamp encoding, and multi-interval composition. A deterministic renderer expresses the same \(\mathcal{Y}\) under the sampled specification. As summarized in \Cref{tab:instruction-response-format-diversity}, this factorization defines 756 possible interface pairings without query paraphrases that could shift the intended evidence. It therefore encourages interface-invariant temporal semantics and format-conditioned serialization, while enabling the controlled \(2\!\times\!2\) analysis in \Cref{tab:instruction-response-format-ablation}.

\begin{center}
\begin{minipage}{\textwidth}
    \centering
    \captionsetup{type=table,hypcap=false}
    \captionof{table}{\textbf{Factorized interface diversity used during SFT.} Independent sampling from 27 grounding requests and 28 response specifications defines 756 possible pairings while the query and target intervals remain fixed. Counts in each response axis partition the same 28 specifications; examples render two intervals.}
    \vspace{-0.7em}
    \label{tab:instruction-response-format-diversity}
    \footnotesize
    \setlength{\tabcolsep}{5pt}
    \renewcommand{\arraystretch}{1.16}
    \begin{tabular}{@{}>{\raggedright\arraybackslash}p{0.17\textwidth}>{\raggedright\arraybackslash}p{0.27\textwidth}>{\centering\arraybackslash}p{0.11\textwidth}>{\raggedright\arraybackslash}p{0.36\textwidth}@{}}
        \toprule
        \textbf{Component} & \textbf{Variation} & \textbf{Count} & \textbf{Representative realization} \\
        \midrule
        \rowcolor{reportbasehighlight}
        Grounding request
        & Question, command, or explicit task schema
        & 27
        & ``When does \(q\) happen?''; ``Locate the segments where \(q\) occurs.''; ``Query: \(q\). Return matching intervals.'' \\
        \rowcolor{reporthighlight}
        Response specification
        & Format-conditioned interval rendering
        & 28
        & Specifies syntax, timestamp encoding, and rules for one or multiple intervals \\
        \addlinespace[0.25em]
        \multicolumn{4}{@{}l}{\textit{Composition of the 28 response specifications}} \\
        \cmidrule(lr){1-4}
        Surface syntax
        & Natural language / labeled fields / bracketed arrays / bare intervals
        & 12 / 5 / 6 / 5
        & \texttt{It happens from 12.0s to 18.5s, and 31.0s to 36.0s.}; \texttt{[[12.0, 18.5], [31.0, 36.0]]} \\
        Timestamp encoding
        & Decimal seconds / minute--second / hour--minute--second
        & 26 / 1 / 1
        & \texttt{12.0s}; \texttt{00:12}; \texttt{00:00:12} \\
        Multi-interval composition
        & Format-consistent joining
        & 28
        & Repeated clauses or interval tokens joined by conjunctions, punctuation, or an outer array \\
        \bottomrule
    \end{tabular}
\end{minipage}
\end{center}

\subsection{Full Temporal Grounding Results}
\label{sec:supp-full-temporal-grounding-results}

The main paper reports R1@0.5 and mIoU; \Cref{tab:temporal-grounding-benchmarks-full} adds R1@0.3 and R1@0.7 wherever available. The progression from loose to strict overlap separates coarse evidence retrieval from boundary precision: a model may find the correct event at R1@0.3 yet fail to delimit it at R1@0.7. Through its 4B and 8B variants, \methodname achieves the highest reported R1@0.3 and R1@0.7 on all seven benchmarks. The 2B variant extends this evaluation to a smaller scale, reaching average R1@0.3, R1@0.5, and R1@0.7 scores of 57.9, 47.0, and 32.0, respectively.

\begin{center}
    \begin{minipage}{\textwidth}
    \centering
    \begingroup
        \colorlet{oursrowfaint}{reporthighlight!55!white}
        \colorlet{oursrowlight}{reporthighlight}
        \colorlet{oursrowdark}{reporthighlightstrong}
        \newcommand{\tgna}{\textcolor{reportslate}{--}}
        \newcommand{\tgsep}{\arrayrulecolor{reportline}\specialrule{0.12pt}{0pt}{0pt}}
        \newcommand{\tgrot}[1]{\rotatebox[origin=c]{0}{#1}}
        \setlength{\tabcolsep}{2pt}
        \renewcommand{\arraystretch}{0.92}
        \fontsize{6.4pt}{6.4pt}\selectfont
        \captionsetup{type=table,hypcap=false}
        \captionof{table}{\textbf{Full temporal grounding results across IoU thresholds.}}
        \vspace{-2mm}
        \label{tab:temporal-grounding-benchmarks-full}
        \resizebox{\textwidth}{!}{%
        \begin{tabular}{@{}>{\raggedright\arraybackslash}p{0.20\textwidth}>{\centering\arraybackslash}p{0.065\textwidth}*{5}{>{\centering\arraybackslash}p{0.091\textwidth}}>{\centering\arraybackslash}p{0.11\textwidth}>{\centering\arraybackslash}p{0.090\textwidth}@{}}
            \toprule
            \textbf{Model} &
            \textbf{Metric} &
            \tgrot{\textbf{Charades}\textsuperscript{TL}} &
            \tgrot{\textbf{ActivityNet}\textsuperscript{TL}} &
            \tgrot{\textbf{QVHighlights}\textsuperscript{TL}} &
            \tgrot{\textbf{VUE-TR}$^{\dag}$} &
            \tgrot{\textbf{VUE-TR-V2}$^{\dag}$} &
            \tgrot{\textbf{MomentSeeker}$^{\ddagger}$} &
            \textbf{\tgrot{Ego4D-NLQ}} \\
            \midrule
            \multirow{4}{*}{GPT-5~\citep{gpt5}} & R1@0.3 & 59.3 & 57.4 & 72.4 & \tgna & 28.0 & \tgna & \tgna \\
            & R1@0.5 & 42.0 & 44.9 & 60.4 & \tgna & 19.5 & \tgna & \tgna \\
            & R1@0.7 & 22.0 & 30.4 & 46.4 & \tgna & 11.6 & \tgna & \tgna \\
            & mIoU & 40.5 & 42.9 & 56.8 & \tgna & 20.0 & \tgna & \tgna \\
            \tgsep
            \multirow{4}{*}{Gemini 3 Pro~\citep{gemini3}} & R1@0.3 & \tgna & \tgna & \tgna & \tgna & 51.5 & \tgna & \tgna \\
            & R1@0.5 & \tgna & \tgna & \tgna & \tgna & 41.3 & \tgna & \tgna \\
            & R1@0.7 & \tgna & \tgna & \tgna & \tgna & 26.0 & \tgna & \tgna \\
            & mIoU & \tgna & \tgna & \tgna & \tgna & 39.7 & \tgna & \tgna \\
            \tgsep
            \multirow{4}{*}{Gemini 2.5 Pro~\citep{Gemini2.5}} & R1@0.3 & 74.1 & 72.3 & 84.1 & 29.1 & \tgna & \tgna & \tgna \\
            & R1@0.5 & 61.1 & 64.2 & 75.9 & 20.5 & \tgna & \tgna & \tgna \\
            & R1@0.7 & 34.0 & 47.1 & 61.1 & 10.3 & \tgna & \tgna & \tgna \\
            & mIoU & 52.8 & 58.1 & 70.4 & 21.9 & \tgna & \tgna & \tgna \\
            \tgsep
            \multirow{4}{*}{Qwen3.5-397B-A17B~\cite{qwen35blog}} & R1@0.3 & 69.3 & 70.6 & 82.1 & 52.4 & 45.6 & 34.3 & 20.5 \\
            & R1@0.5 & 47.8 & 59.1 & 71.6 & 42.7 & 35.5 & 22.3 & 13.3 \\
            & R1@0.7 & 24.9 & 41.7 & 56.1 & 33.1 & 25.5 & 9.9 & 7.1 \\
            & mIoU & 47.5 & 53.8 & 65.8 & 42.2 & 34.5 & 23.3 & 14.5 \\
            \tgsep
            \multirow{4}{*}{Qwen3-VL-235B-A22B~\citep{qwen3vl}} & R1@0.3 & 71.7 & 69.0 & 79.6 & 55.8 & 44.1 & 34.4 & 17.8 \\
            & R1@0.5 & 50.8 & 57.5 & 70.2 & 42.5 & 34.5 & 21.2 & 11.5 \\
            & R1@0.7 & 24.5 & 39.3 & 54.5 & 31.6 & 24.1 & 11.3 & 6.3 \\
            & mIoU & 47.8 & 52.2 & 64.6 & 43.1 & 34.1 & 23.7 & 12.7 \\
            \tgsep
            \multirow{4}{*}{Qwen3.5-35B-A3B~\cite{qwen35blog}} 
            & R1@0.3 & 69.5 & 69.4 & 81.6 & 56.2 & 47.0 & 33.2 & 18.2 \\
            & R1@0.5 & 50.4 & 58.6 & 72.6 & 46.3 & 37.1 & 20.2 & 12.0 \\
            & R1@0.7 & 27.3 & 40.1 & 56.5 & 32.4 & 24.9 & 9.7 & 6.9 \\
            & mIoU & 48.2 & 52.6 & 66.0 & 44.6 & 35.3 & 22.6 & 13.0 \\
            \tgsep
            \multirow{4}{*}{Qwen3-VL-30B-A3B~\citep{qwen3vl}} & R1@0.3 & 70.3 & 65.7 & 79.3 & 53.0 & 42.7 & 33.6 & 16.1 \\
            & R1@0.5 & 46.5 & 51.1 & 67.6 & 40.4 & 32.3 & 19.6 & 10.4 \\
            & R1@0.7 & 25.1 & 36.5 & 52.6 & 32.2 & 21.8 & 9.2 & 5.5 \\
            & mIoU & 48.1 & 49.4 & 63.2 & 42.5 & 32.6 & 22.6 & 11.5 \\
            \tgsep
            \multirow{4}{*}{Vidi-1.5-9B~\citep{vidi}} & R1@0.3 & 28.7 & 51.8 & 68.7 & 57.3 & 43.9 & 27.0 & 11.4 \\
            & R1@0.5 & 18.3 & 40.2 & 58.4 & 45.0 & 31.9 & 14.8 & 6.5 \\
            & R1@0.7 & 10.1 & 28.2 & 46.2 & 36.0 & 22.2 & 6.9 & 3.4 \\
            & mIoU & 22.0 & 39.3 & 55.9 & 47.0 & 34.3 & 18.6 & 8.4 \\
            \tgsep
            \multirow{4}{*}{LLaVA-OneVision-2-8B~\citep{LLaVA-OneVision-2}} & R1@0.3 & 73.1 & 65.6 & 78.2 & 51.2 & 46.3 & 28.8 & 18.2 \\
            & R1@0.5 & 59.6 & 57.9 & 70.0 & 40.8 & 35.0 & 18.0 & 12.0 \\
            & R1@0.7 & 34.0 & 40.9 & 57.0 & 30.1 & 22.8 & 8.2 & 6.4 \\
            & mIoU & 52.6 & 52.4 & 65.7 & 41.3 & 34.5 & 19.6 & 12.8 \\
            \tgsep
            \multirow{4}{*}{Qwen3-VL-8B~\citep{qwen3vl}} & R1@0.3 & 68.5 & 63.8 & 74.4 & 33.0 & 19.6 & 13.6 & 7.4 \\
            & R1@0.5 & 49.3 & 51.1 & 62.9 & 23.8 & 13.1 & 9.4 & 4.8 \\
            & R1@0.7 & 26.2 & 35.0 & 49.3 & 17.5 & 8.3 & 4.5 & 2.5 \\
            & mIoU & 47.2 & 48.0 & 59.4 & 25.5 & 14.0 & 9.8 & 5.3 \\
            \tgsep
            \multirow{4}{*}{InternVideo3-8B~\citep{InternVideo3}} & R1@0.3 & 75.8 & 60.0 & 72.2 & 51.0 & 40.8 & 28.8 & 12.3 \\
            & R1@0.5 & 61.7 & 51.4 & 62.6 & 39.4 & 32.4 & 17.9 & 7.3 \\
            & R1@0.7 & 32.8 & 33.4 & 48.9 & 32.0 & 21.7 & 7.3 & 3.2 \\
            & mIoU & 53.2 & 46.3 & 59.2 & 41.1 & 32.6 & 19.3 & 8.6 \\
            \tgsep
            \multirow{4}{*}{TimeLens-8B~\citep{zhang2026timelens}} & R1@0.3 & 78.8 & 71.8 & 81.2 & 56.4 & 46.2 & 32.0 & 21.9 \\
            & R1@0.5 & 65.8 & 62.0 & 71.7 & 42.9 & 34.4 & 20.4 & 14.6 \\
            & R1@0.7 & 37.6 & 43.8 & 57.1 & 31.2 & 23.5 & 9.9 & 7.7 \\
            & mIoU & 57.0 & 56.3 & 66.8 & 43.6 & 34.0 & 21.9 & 15.7 \\
            \tgsep
            \multirow{4}{*}{Video-o3-7B~\citep{Video-o3}} & R1@0.3 & 58.7 & 53.2 & 62.7 & 32.8 & 21.4 & 18.2 & 4.2 \\
            & R1@0.5 & 34.2 & 41.0 & 49.5 & 22.1 & 12.9 & 9.8 & 1.9 \\
            & R1@0.7 & 16.2 & 24.4 & 34.4 & 14.5 & 7.0 & 3.9 & 0.7 \\
            & mIoU & 38.6 & 38.9 & 47.4 & 25.1 & 15.3 & 13.0 & 3.1 \\
            \tgsep
            \multirow{4}{*}{VideoChat-Flash-7B~\citep{videochat-flash}} & R1@0.3 & 60.2 & 35.5 & 45.2 & 20.2 & 13.7 & 10.9 & 3.1 \\
            & R1@0.5 & 37.9 & 21.8 & 30.6 & 15.4 & 9.5 & 5.9 & 1.5 \\
            & R1@0.7 & 17.8 & 10.5 & 16.7 & 10.7 & 5.6 & 2.2 & 0.6 \\
            & mIoU & 39.7 & 24.8 & 32.7 & 17.3 & 10.1 & 7.2 & 2.5 \\
            \tgsep
            \multirow{4}{*}{MiMo-VL-7B~\citep{MiMo-VL}} & R1@0.3 & 57.9 & 49.3 & 57.1 & 25.1 & 14.7 & 6.5 & 0.9 \\
            & R1@0.5 & 42.6 & 38.7 & 42.6 & 17.5 & 10.1 & 3.8 & 0.4 \\
            & R1@0.7 & 20.5 & 22.4 & 28.4 & 11.2 & 4.8 & 1.6 & 0.1 \\
            & mIoU & 39.6 & 35.5 & 41.5 & 19.1 & 10.4 & 6.0 & 1.0 \\
            \tgsep
            \multirow{4}{*}{TimeSuite-7B~\citep{timesuite}} & R1@0.3 & 56.3 & 27.1 & 27.1 & 15.8 & 9.4 & 6.9 & 1.6 \\
            & R1@0.5 & 35.5 & 17.5 & 16.9 & 10.3 & 5.3 & 3.3 & 0.6 \\
            & R1@0.7 & 18.0 & 8.6 & 9.9 & 7.0 & 3.2 & 1.2 & 0.3 \\
            & mIoU & 38.1 & 19.8 & 21.7 & 13.2 & 7.3 & 5.9 & 1.4 \\
            \tgsep
            \multirow{4}{*}{InternVL3.5-4B~\citep{internvl3.5}} & R1@0.3 & 24.6 & 20.0 & 25.5 & 20.4 & 11.0 & 3.7 & 2.0 \\
            & R1@0.5 & 14.1 & 12.7 & 15.8 & 14.1 & 6.0 & 2.1 & 0.7 \\
            & R1@0.7 & 6.2 & 8.2 & 6.2 & 9.5 & 3.5 & 0.6 & 0.1 \\
            & mIoU & 16.0 & 14.9 & 17.7 & 16.5 & 8.5 & 3.2 & 2.0 \\
            \tgsep
            \multirow{4}{*}{Qwen3-VL-4B~\citep{qwen3vl}} & R1@0.3 & 71.8 & 67.4 & 78.4 & 41.0 & 25.6 & 21.5 & 15.5 \\
            & R1@0.5 & 53.8 & 54.8 & 66.4 & 33.0 & 19.9 & 13.5 & 10.6 \\
            & R1@0.7 & 28.7 & 37.6 & 51.1 & 25.0 & 13.3 & 5.4 & 6.0 \\
            & mIoU & 49.4 & 50.7 & 62.3 & 33.6 & 20.3 & 15.3 & 11.4 \\
            \tgsep
            \multirow{4}{*}{Molmo2-4B~\citep{Molmo2}} & R1@0.3 & 44.4 & 50.8 & 73.7 & 52.0 & 41.7 & 25.2 & 11.8 \\
            & R1@0.5 & 31.1 & 40.3 & 62.6 & 42.5 & 30.9 & 14.7 & 7.7 \\
            & R1@0.7 & 18.5 & 29.9 & 51.8 & 33.7 & 20.8 & 8.0 & 4.5 \\
            & mIoU & 34.7 & 40.9 & 60.8 & 43.1 & 32.1 & 19.5 & 10.0 \\
            \tgsep
            \multirow{4}{*}{VideoChat3-4B} & R1@0.3 & 78.4 & 70.1 & 81.0 & 61.0 & 53.8 & 37.2 & 20.9 \\
            & R1@0.5 & 64.9 & 60.5 & 72.5 & 47.8 & 40.4 & 24.4 & 13.9 \\
            & R1@0.7 & 35.9 & 42.2 & 56.8 & 34.7 & 27.3 & 12.0 & 7.2 \\
            & mIoU & 56.1 & 54.6 & 67.0 & 47.9 & 40.2 & 25.9 & 14.6 \\
            \tgsep
            \multirow{4}{*}{Marlin-2B~\citep{nemostation2026marlin2b}} & R1@0.3 & 66.3 & 50.5 & 60.7 & 31.6 & 24.5 & 23.9 & 12.7 \\
            & R1@0.5 & 51.7 & 40.5 & 48.5 & 23.6 & 19.9 & 15.2 & 7.8 \\
            & R1@0.7 & 27.1 & 25.3 & 34.9 & 16.4 & 12.0 & 5.8 & 3.8 \\
            & mIoU & 46.5 & 37.9 & 46.8 & 24.8 & 19.2 & 16.3 & 8.9 \\
            \tgsep
            \multirow{4}{*}{Qwen3.5-2B~\citep{qwen35blog}} & R1@0.3 & 67.3 & 64.5 & 74.8 & 34.7 & 28.7 & 22.2 & 15.9 \\
            & R1@0.5 & 46.8 & 52.8 & 65.7 & 27.2 & 21.5 & 12.8 & 10.1 \\
            & R1@0.7 & 24.5 & 35.9 & 50.5 & 17.0 & 13.7 & 6.1 & 5.1 \\
            & mIoU & 46.0 & 48.9 & 60.6 & 27.8 & 21.6 & 15.6 & 11.2 \\
            \tgsep
            \multirow{4}{*}{Qwen3-VL-2B~\citep{qwen3vl}} & R1@0.3 & 62.7 & 53.7 & 67.8 & 41.9 & 33.7 & 17.3 & 13.3 \\
            & R1@0.5 & 44.4 & 41.0 & 53.4 & 29.9 & 22.5 & 9.8 & 8.1 \\
            & R1@0.7 & 23.4 & 26.5 & 39.3 & 18.5 & 13.8 & 3.2 & 3.8 \\
            & mIoU & 43.4 & 39.9 & 52.2 & 31.4 & 24.1 & 12.4 & 9.3 \\
            \tgsep
            \rowcolor{oursrowfaint} & R1@0.3 & 74.2 & 71.1 & 81.1 & 64.2 & 57.5 & 34.1 & 23.3 \\
            \rowcolor{oursrowfaint} & R1@0.5 & 60.6 & 62.0 & 72.8 & 50.7 & 44.5 & 21.9 & 16.5 \\
            \rowcolor{oursrowfaint} & R1@0.7 & 33.6 & 43.2 & 57.8 & 38.9 & 31.5 & 9.8 & 9.1 \\
            \rowcolor{oursrowfaint}\multirow{-4}{*}{\methodname-2B} & mIoU & 53.4 & 55.4 & 67.0 & 51.1 & 43.8 & 24.3 & 16.8 \\
            \tgsep
            \rowcolor{oursrowlight} & R1@0.3 & 80.3 & 74.6 & 83.3 & 66.5 & 62.9 & 38.8 & 26.0 \\
            \rowcolor{oursrowlight} & R1@0.5 & 66.6 & 65.5 & 75.2 & 52.4 & 49.5 & 27.0 & 18.1 \\
            \rowcolor{oursrowlight} & R1@0.7 & 37.5 & 47.2 & 60.0 & 41.1 & 35.5 & 12.5 & 10.0 \\
            \rowcolor{oursrowlight}\multirow{-4}{*}{\methodname-4B} & mIoU & 57.7 & 59.0 & 69.3 & 53.2 & 48.1 & 27.9 & 18.6 \\
            \tgsep
            \rowcolor{oursrowdark} & R1@0.3 & 80.9 & 74.0 & 84.4 & 65.3 & 61.8 & 40.8 & 27.0 \\
            \rowcolor{oursrowdark} & R1@0.5 & 68.0 & 65.6 & 76.1 & 51.6 & 49.1 & 27.3 & 18.4 \\
            \rowcolor{oursrowdark} & R1@0.7 & 38.9 & 46.6 & 61.2 & 41.9 & 34.8 & 13.7 & 9.6 \\
            \rowcolor{oursrowdark}\multirow{-4}{*}{\methodname-8B} & mIoU & 58.6 & 58.6 & 70.2 & 53.5 & 47.7 & 28.5 & 19.0 \\
            \bottomrule
        \end{tabular}%
        }
    \endgroup
    \end{minipage}
    \end{center}

\clearpage

\subsection{Additional Qualitative Examples}
\label{sec:supp-qualitative-examples}

\begin{center}
    \includegraphics[width=\textwidth]{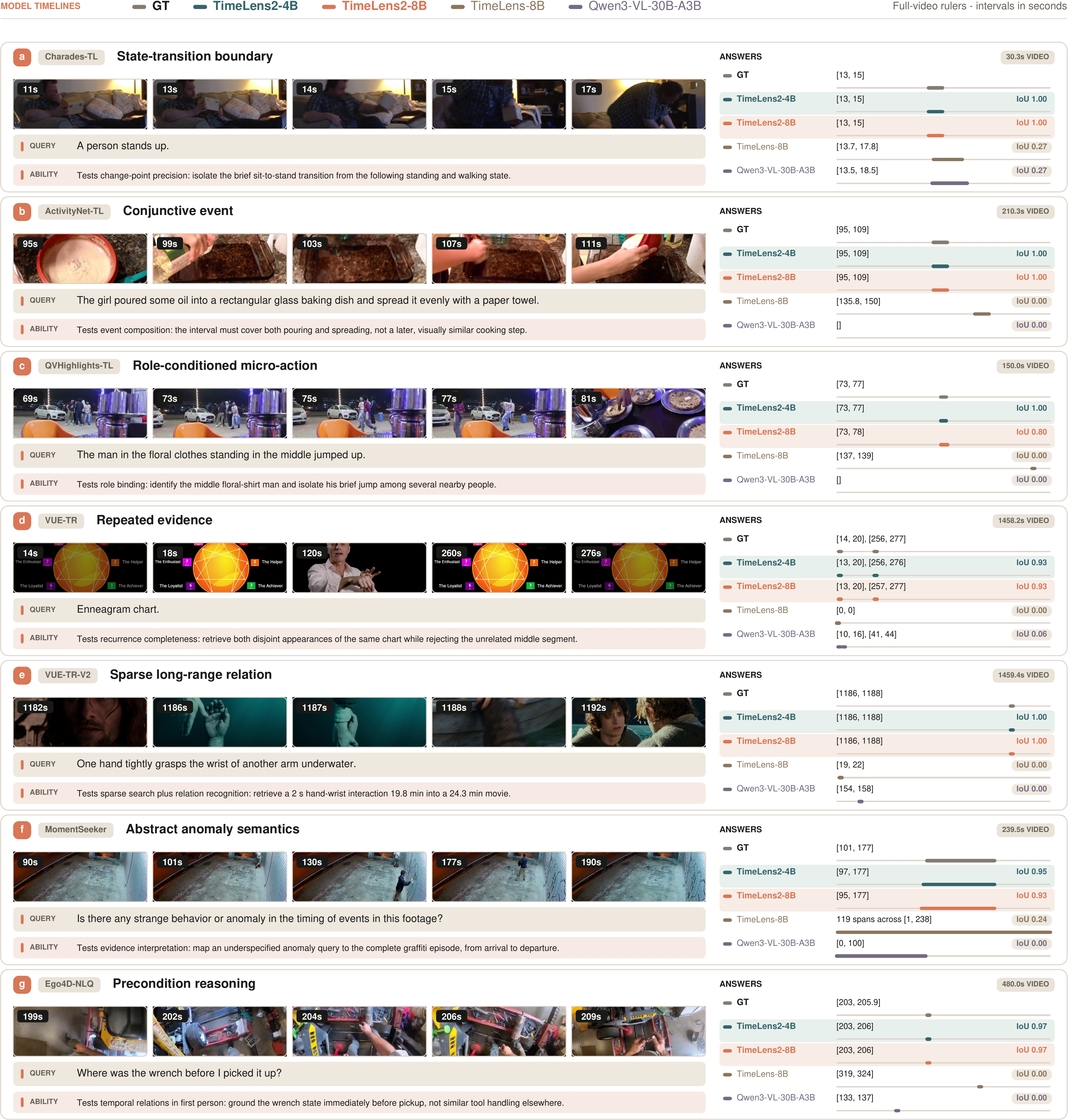}
    \captionof{figure}{\textbf{Additional qualitative results on all seven benchmarks.} Each row pairs five frames near the target evidence with the query, the temporal reasoning challenge, and full-video timelines showing the ground truth and predictions from four models. The \methodname variants consistently recover the intended evidence, whereas the baselines select distractors, miss events, overextend boundaries, or fragment their predictions.}
    \label{fig:supp-qualitative-examples}
\end{center}

These examples probe complementary temporal-reasoning capabilities. Charades-TL tests precise state-transition boundaries; ActivityNet-TL requires one interval covering both actions in a conjunction; QVHighlights-TL tests role-conditioned entity binding; and VUE-TR requires complete retrieval of recurring evidence. VUE-TR-V2 adds sparse search for a two-second relation nearly 20 minutes into a film. MomentSeeker requires mapping an underspecified anomaly query to the complete graffiti episode, rather than broad or fragmented coverage. Ego4D-NLQ tests precondition reasoning by locating the wrench immediately before pickup. Across all settings, \methodname remains precise, complete, and resistant to visually similar distractors.

\FloatBarrier
\clearpage

\subsection{Temporal Grounding Evaluation Protocol}
\label{sec:supp-evaluation-protocol}

\paragraph{Benchmarks and evaluation subsets.}
We conduct all in-house evaluations with VLMEvalKit~\citep{VLMEvalKit}.  Within this framework, we implement evaluation adapters for the TimeLens-Bench re-annotations of Charades-STA, ActivityNet Captions, and QVHighlights; the vision-query subsets of VUE-TR and VUE-TR-V2; the text-only query subset of MomentSeeker; and the official Ego4D-NLQ-v2 validation split. We use visual frames only: audio and subtitles are not supplied to the model.

\paragraph{Standard video input configuration.}
For models evaluated, the input configuration is fixed per benchmark rather than tuned per model; the exact constructor arguments are given in \Cref{tab:supp-standard-eval-settings}.  The video processor receives \texttt{min\_pixels} and \texttt{max\_pixels} as per-frame pixel-area bounds and \texttt{total\_pixels} as the aggregate visual pixel budget.  We first sample at the specified frame rate.  If \(\lfloor T\!\times\!\mathrm{fps}\rfloor\) exceeds \texttt{frames\_limit}, we uniformly retain the capped number of frames over the entire duration \(T\), and pass the resulting effective sampling rate to the model.
\begin{table*}[!b]
    \centering
    \caption{\textbf{Standard input configuration.} Exact arguments for the seven temporal grounding benchmarks.}
    \label{tab:supp-standard-eval-settings}
    \fontsize{8.2pt}{9.6pt}\selectfont
    \setlength{\tabcolsep}{4.2pt}
    \renewcommand{\arraystretch}{1}
    \begin{tabular}{@{}lccccccc@{}}
        \toprule
        \multirow{2}{*}{\textbf{Benchmark}} &
        \multirow{2}{*}{\shortstack{\textbf{Queries}\\[-0.25ex]{\scriptsize\color{reportslate}\(N\)}}} &
        \multicolumn{2}{c}{\reporttablegroup{Temporal sampling}} &
        \multicolumn{2}{c}{\reporttablegroup{Per-frame area}} &
        \multirow{2}{*}{\shortstack{\reporttablegroup{Total area}\\[-0.25ex]{\scriptsize\texttt{total\_pixels}}}} &
        \multirow{2}{*}{\textbf{Output}} \\
        \cmidrule(lr){3-4}\cmidrule(lr){5-6}
        & & \textbf{FPS} & \shortstack{\scriptsize\texttt{frames\_}\\[-0.3ex]\scriptsize\texttt{limit}} &
        \shortstack{\scriptsize\texttt{min\_}\\[-0.3ex]\scriptsize\texttt{pixels}} &
        \shortstack{\scriptsize\texttt{max\_}\\[-0.3ex]\scriptsize\texttt{pixels}} & & \\
        \midrule
        \rowcolor{reporthighlight!55}Charades-TimeLens & 3,363 & 4 & 2,048 & \(32^2\) & \(640^2\) & \(128000\!\times\!32^2\) & one interval \\
        \rowcolor{reporthighlight!55}ActivityNet-TimeLens & 4,500 & 4 & 2,048 & \(32^2\) & \(640^2\) & \(128000\!\times\!32^2\) & one interval \\
        \rowcolor{reporthighlight!55}QVHighlights-TimeLens & 1,541 & 4 & 2,048 & \(32^2\) & \(640^2\) & \(128000\!\times\!32^2\) & one interval \\
        \addlinespace[0.15em]
        \rowcolor{reportbasehighlight!58}VUE-TR (vision) & 525 & 1 & 2,048 & \(32^2\) & \(480^2\) & \(128000\!\times\!32^2\) & all intervals \\
        \rowcolor{reportbasehighlight!58}VUE-TR-V2 (vision) & 715 & 1 & 2,048 & \(32^2\) & \(480^2\) & \(128000\!\times\!32^2\) & all intervals \\
        \addlinespace[0.15em]
        MomentSeeker (text-only) & 1,000 & 2 & 2,048 & \(32^2\) & \(480^2\) & \(128000\!\times\!32^2\) & all intervals \\
        Ego4D-NLQ-v2 (val) & 4,552 & 2 & 2,048 & \(32^2\) & \(480^2\) & \(128000\!\times\!32^2\) & one interval \\
        \bottomrule
    \end{tabular}
\end{table*}

\paragraph{Model-specific input settings.}
All evaluated Qwen3-VL and Qwen3.5 variants, LLaVA-OneVision-2-8B, TimeLens-8B, VideoChat3-4B, InternVideo3-8B, Molmo2-4B, and \methodname-2B/4B/8B use the standard per-benchmark settings in \Cref{tab:supp-standard-eval-settings}.  Every deviation is summarized in \Cref{tab:supp-model-eval-deviations} by model and dataset group.  These deviations are used when a model cannot accommodate the standard context or frame budget.
All Qwen3.5 models except Qwen3.5-2B are evaluated with thinking enabled, which substantially improves temporal-grounding performance.  For Qwen3.5-2B, we disable thinking because it frequently enters non-terminating reasoning loops, especially on long-video benchmarks.  Thinking is disabled for every other evaluated model.
Because Molmo2 follows generic instructions and prescribed output formats unreliably, we adopt the native spatiotemporal-grounding question template from its original work.  Specifically, we use the prompt ``Point the start and end of the event described by the sentence \emph{query}.'' and parse the first two temporal coordinates from its native \texttt{<points coords="...">} or \texttt{<tracks coords="...">} response.

\begin{table*}[t]
    \centering
    \caption{\textbf{Model-specific overrides.} Input configurations that deviate from \Cref{tab:supp-standard-eval-settings}. TL denotes the three TimeLens-Bench datasets.}
    \label{tab:supp-model-eval-deviations}
    \fontsize{8.1pt}{9.5pt}\selectfont
    \setlength{\tabcolsep}{1pt}
    \renewcommand{\arraystretch}{1}
    \begin{tabular}{@{}>{\raggedright\arraybackslash}p{0.22\textwidth}>{\raggedright\arraybackslash}p{0.21\textwidth}>{\centering\arraybackslash}p{0.06\textwidth}>{\centering\arraybackslash}p{0.09\textwidth}>{\centering\arraybackslash}p{0.115\textwidth}>{\centering\arraybackslash}p{0.115\textwidth}>{\centering\arraybackslash}p{0.155\textwidth}@{}}
        \toprule
        \textbf{Model} & \textbf{Dataset group} & \textbf{FPS} &
        \shortstack{\textbf{Frame}\\[-0.25ex]\textbf{cap}} &
        \shortstack{\textbf{Min.}\\[-0.25ex]{\scriptsize\texttt{min\_pixels}}} &
        \shortstack{\textbf{Max.}\\[-0.25ex]{\scriptsize\texttt{max\_pixels}}} &
        \shortstack{\textbf{Total area}\\[-0.25ex]{\scriptsize\texttt{total\_pixels}}} \\
        \midrule
        \multicolumn{7}{@{}l}{\reporttablegroup{Global overrides}} \\
        \addlinespace[0.05em]
        \rowcolor{reportbasehighlight!58}Video-o3-7B / MiMo-VL-7B & All seven & 2 & 768 & \(10\!\times\!28^2\) & \(768\!\times\!28^2\) & \(24576\!\times\!28^2\) \\
        \rowcolor{reportbasehighlight!58}VideoChat-Flash-7B & All seven & \(\sim\!1\) & 512 & \(448^2\) & \(448^2\) & \reportna \\
        \rowcolor{reportbasehighlight!58}TimeSuite-7B & All seven & \reportna & 128 & \(224^2\) & \(224^2\) & \reportna \\
        \addlinespace[0.28em]
        \multicolumn{7}{@{}l}{\reporttablegroup{Dataset-specific override}} \\
        \addlinespace[0.05em]
        \rowcolor{reporthighlight!55}InternVL3.5-4B & TL & 1 & 32 & \(28^2\) & \(448^2\) & \(12800\!\times\!32^2\) \\
        \rowcolor{reporthighlight!55} & VUE-TR & 1 & 128 & \(28^2\) & \(448^2\) & \(12800\!\times\!32^2\) \\
        \rowcolor{reporthighlight!55} & VUE-TR-V2 / MomentSeeker & 1 & 128 & \(32^2\) & \(480^2\) & \(12800\!\times\!32^2\) \\
        \rowcolor{reporthighlight!55} & Ego4D-NLQ & 1 & 64 & \(32^2\) & \(480^2\) & \(12800\!\times\!32^2\) \\
        \bottomrule
    \end{tabular}
\end{table*}

\paragraph{Prompt templates.}
The evaluation entry point uses the dataset-side prompts in \Cref{tab:supp-prompt-templates}.

\begin{table*}[t]
    \centering
    \caption{\textbf{Dataset-side prompt templates.} Exact templates used in the standard temporal grounding evaluation; \texttt{\{query\}} is replaced by the benchmark query.}
    \label{tab:supp-prompt-templates}
    \fontsize{7.7pt}{9.2pt}\selectfont
    \setlength{\tabcolsep}{5.5pt}
    \renewcommand{\arraystretch}{0.95}
    \begin{tabular}{@{}>{\raggedright\arraybackslash}p{0.205\textwidth}>{\raggedright\arraybackslash\ttfamily}p{0.74\textwidth}@{}}
        \toprule
        \textbf{Benchmark group} & \textnormal{\textbf{Prompt template}} \\
        \midrule
        \rowcolor{reporthighlight!55}\textnormal{Charades-/ActivityNet-/ QVHighlights-TimeLens} &
        Please find the visual event described by the sentence '\{query\}', determining its starting and ending times. The format should be: 'The event happens in \textless start time\textgreater{} - \textless end time\textgreater{} seconds'. \\
        \rowcolor{reportbasehighlight!58}\textnormal{VUE-TR / VUE-TR-V2} &
        You are given a video.\newline
        Task: temporal retrieval.\newline
        Given the query: "\{query\}", return ALL time spans (in seconds) where the query is relevant.\newline
        Output format MUST be a JSON array of [start, end] pairs, e.g. [[0, 3.5], [10, 12]]. \\
        \textnormal{MomentSeeker} &
        You are given a video.\newline
        Task: temporal grounding.\newline
        Given the query: "\{query\}", return ALL time spans (in seconds) where the query is grounded in the video.\newline
        Output format MUST be a JSON array of [start, end] pairs, e.g. [[0, 3.5], [10, 12]]. \\
        \rowcolor{reportbasehighlight!58}\textnormal{Ego4D-NLQ-v2 (val)} &
        You are given a video.\newline
        Task: temporal grounding.\newline
        Given the query: "\{query\}", return the time span (in seconds) where the query is grounded in the video.\newline
        Output format MUST be [[start, end]]. \\
        \bottomrule
    \end{tabular}
\end{table*}

\subsection{Train--Test Overlap Audit}
\label{sec:supp-train-test-overlap}

We canonicalized the complete list of training-source YouTube IDs and compared it with the video identifiers used by all seven evaluation subsets. Specifically, we removed the \texttt{v\_} prefix from ActivityNet IDs, reduced QVHighlights clip names to their 11-character source IDs, and used the native YouTube IDs provided by VUE-TR, VUE-TR-V2, and the explicitly identified YouTube videos in MomentSeeker. We additionally searched each benchmark's \texttt{video}, \texttt{video\_id}, and \texttt{src\_video\_path} fields for embedded training IDs. The intersection was \emph{empty} for every benchmark. Charades and Ego4D use non-YouTube identifiers, while most MomentSeeker videos are distributed under renamed internal identifiers without a source-ID mapping; thus, this audit establishes \emph{zero} overlap among recoverable source identifiers but cannot exclude content duplicates concealed by renaming.

\subsection{Limitations and Future Directions}
\label{sec:supp-limitations-future-work}

\paragraph{Limitations.}
TimeLens2-93K was constructed under finite financial and computational budgets.  We therefore relied on cost-effective, predominantly open-weight models throughout the pipeline: Qwen3-VL-235B-A22B-Instruct for captioning, Kimi-K2.5 for query and proposal generation, Qwen3-VL-30B-A3B and TimeLens-8B for temporal annotation, and Qwen3-VL-235B-A22B-Instruct for final refinement.  Although this pipeline already produces effective training data, its quality is inevitably bounded by the capabilities of these models.  Replacing them with stronger proprietary multimodal models, such as Gemini 3.1 Pro~\citep{gemini31pro}, could substantially improve the richness, diversity, and temporal precision of the resulting annotations.  We therefore believe that the current dataset is not the endpoint of this recipe, but a first demonstration with considerable room to scale.

\paragraph{Future directions.}
Beyond improving temporal grounding data with stronger models, our construction framework may naturally extend to long-context spatiotemporal grounding, where a model must determine not only \emph{when} an event occurs, but also \emph{where} relevant entities appear and how they evolve over time.  This capability is particularly important for embodied intelligence: an agent operating over long horizons must connect past observations with the current scene, track changes in objects and environments, and ground its decisions in concrete visual evidence before completing downstream tasks.  We hope our work can inspire larger and richer spatiotemporally grounded datasets, as well as models that turn long-video perception into persistent memory, grounded reasoning, and ultimately reliable action.

\end{document}